\documentclass[10pt,twocolumn,letterpaper]{article}

\newcommand{\ie}{\textit{i}.\textit{e}.,~}
\newcommand{\eg}{\textit{e}.\textit{g}.,~}
\usepackage{wacv}
\usepackage{times}
\usepackage{epsfig}
\usepackage{graphicx}
\usepackage{amsmath}
\usepackage{amssymb}
\usepackage{graphbox}
\usepackage{pifont}
\usepackage[export]{adjustbox}
\usepackage{multirow}
\usepackage{makecell}
\usepackage{booktabs}
\usepackage[accsupp]{axessibility}
\usepackage[symbol]{footmisc} 
\def\wacvPaperID{179} % Enter the WACV Paper ID here

\wacvalgorithmstrack   % Uncomment this line if you are submitting to the Algorithms Track.
\wacvfinalcopy % *** Uncomment this line for the final submission

\ifwacvfinal
\usepackage[breaklinks=true,bookmarks=false]{hyperref}
\else
\usepackage[pagebackref=true,breaklinks=true,colorlinks,bookmarks=false]{hyperref}
\fi

\begin{document}

%%%%%%%%% TITLE
\title{IFQA: Interpretable Face Quality Assessment}
%\footnote[3]{}
\author{Byungho Jo$^{1}$\quad Donghyeon Cho$^{2}$\quad In Kyu Park$^{1}$\quad Sungeun Hong$^{1}$\\
$^{1}$Inha University\quad$^{2}$Chungnam National University\\
{\tt\small byunghojo12@gmail.com}~~{\tt\small cdh12242@cnu.ac.kr} ~~{\tt\small \{pik, csehong\}@inha.ac.kr}}
% \and
% Donghyeon Cho\\
% Chungnam National University\\

% \and
% In Kyu Park, Sungeun Hong\\
% Inha University\\

% \and
% Sungeun Hong\\
% Inha University\\

\maketitle
\thispagestyle{empty}

%%%%%%%%% ABSTRACT
\begin{abstract}
Existing face restoration models have relied on general assessment metrics that do not consider the characteristics of facial regions.
% , resulting in a low correlation with human visual perception. 
Recent works have therefore assessed their methods using human studies, which is not scalable and involves significant effort. This paper proposes a novel face-centric metric based on an adversarial framework where a generator simulates face restoration and a discriminator assesses image quality.
Specifically, our per-pixel discriminator enables interpretable evaluation that cannot be provided by traditional metrics. 
Moreover, our metric emphasizes facial primary regions considering that even minor changes to the eyes, nose, and mouth significantly affect human cognition. 
Our face-oriented metric consistently surpasses existing general or facial image quality assessment metrics by impressive margins. 
We demonstrate the generalizability of the proposed strategy in various architectural designs and challenging scenarios.
Interestingly, we find that our IFQA can lead to performance improvement as an objective function.
The code and models are available at \url{https://github.com/VCLLab/IFQA}.
\end{abstract}
%\footnote[3]{Corresponding author.}
\vspace{-2mm}
%%%%%%%%% BODY TEXT
\section{Introduction}

\begin{figure} 
    \begin{tabular}{l@{\hskip 0.8pt}c@{\hskip 1pt}c|c@{\hskip 1pt}c}
\multicolumn{3}{c}{\textbf{w/ Reference}} & \multicolumn{2}{c}{\textbf{w/o Reference}}\\\toprule
\centering
\small{Reference}& \small{Image A}& \small{Image B} & \small{Image A} & \small{Image B}\\
\includegraphics[width=0.16\linewidth]{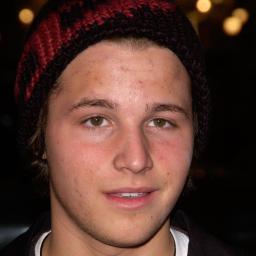}&
\includegraphics[width=0.16\linewidth]{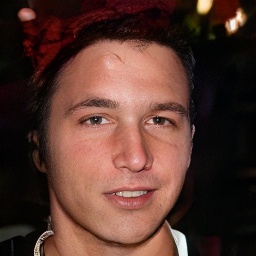}& 
\includegraphics[width=0.16\linewidth]{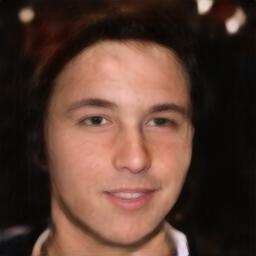}& 
\includegraphics[width=0.16\linewidth]{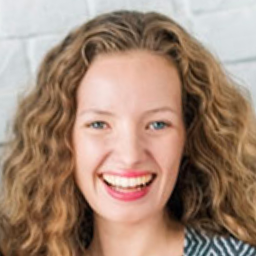}&
\includegraphics[width=0.16\linewidth]{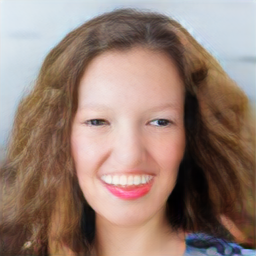}\\
\includegraphics[width=0.16\linewidth]{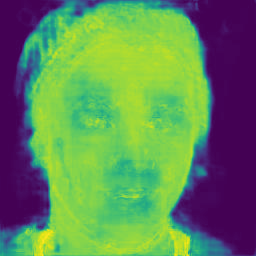}&
\includegraphics[width=0.16\linewidth]{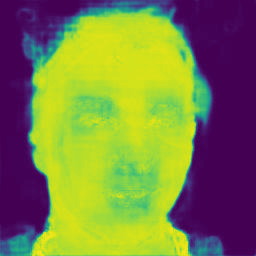}&
\includegraphics[width=0.16\linewidth]{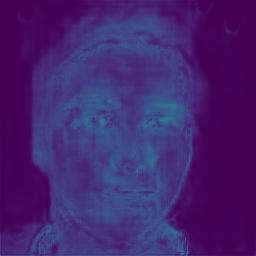}&
\includegraphics[width=0.16\linewidth]{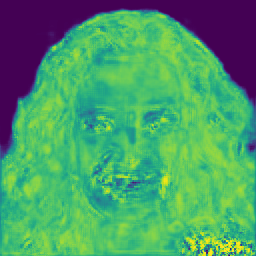}&
\includegraphics[width=0.16\linewidth]{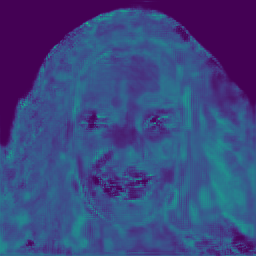}\\
\scriptsize{{PSNR~\cite{PSNR}}}& & \color{green}\ding{52}& \footnotesize{\textbf{N/A}}&\\ % 
\scriptsize{{SSIM~\cite{SSIM}}}& & \color{green}\ding{52}& \footnotesize{\textbf{N/A}}&\\ % 
\scriptsize{{LPIPS~\cite{LPIPS}}}& \color{green}\ding{52}& & \footnotesize{\textbf{N/A}}&\\
\scriptsize{{NIQE~\cite{NIQE}}}& & \color{red}\ding{52}& &\color{red}\ding{52}\\
\scriptsize{{BRISQUE~\cite{Brisque}}}& \color{red}\ding{52}& & &\color{red}\ding{52}\\ %
\scriptsize{{PI~\cite{srmetricPI}}}& \color{red}\ding{52}& & &\color{red}\ding{52}\\ %
\scriptsize{{FIQA~\cite{Ortega19, faceqnetv1, SER-FIQ, SDD-FIQA}}}& \color{red}\ding{52}& & &\color{red}\ding{52}\\ %
\scriptsize{{IFQA (Ours)}}& \color{red}\ding{52}& & \color{red}\ding{52}&\\ %
\scriptsize{{Human}}& \ding{52}& & \ding{52}&\\
  \bottomrule
     \multicolumn{5}{c}{\small{Human judgment: \ding{52} \quad Full-Ref. IQA: \color{green}\ding{52} \quad   \color{black} No-Ref. IQA: \color{red}\ding{52}}}  \\ \\
    \end{tabular} 
    \caption{
    Which of `Image A' or `Image B' is closer to the given reference image or looks high-quality?
    General full-reference metrics (\eg PSNR/SSIM), no-reference metrics (\eg NIQE, BRISQUE, PI), and FIQA methods are inconsistent with human judgment. 
    LPIPS agrees with human judgment but cannot be applied to the blind face restoration scenario.
    Our IFQA is consistent with human judgment and can provide interpretability maps where the brighter the area, the higher the quality.}
    % as shown in the second row.}
                % \vspace{-1.5mm}
    \label{fig:teaser}
\end{figure}
% Most face-related studies (\eg face detection~\cite{deng2020retinaface}, face identification~\cite{deng2019arcface}, and face manipulation~\cite{deefake}) have targeted face images taken in a relatively stable environment.
% However, face images are often degraded in the real world owing to factors such as blur, poor illumination, and low resolution.
Considerable efforts have been devoted to restoring facial images from degraded images~\cite{PULSE, GPEN, GFP-GAN}. 
Conventional face restoration studies adopt full-reference metrics widely used in general image restoration, \eg PSNR~\cite{PSNR}, SSIM~\cite{SSIM}, LPIPS~\cite{LPIPS}, to evaluate the similarity between reference and restored images.
Blind face restoration (BFR) studies that can handle multiple unknown degradations adopt no-reference metrics such as NIQE~\cite{NIQE} and BRISQUE~\cite{Brisque}.
However, because existing general metrics do not consider facial characteristics, their judgments could differ from human perceptions as shown in Figure~\ref{fig:teaser}. 

Recent face restoration studies~\cite{Wan_OldPhoto, PULSE, GPEN} have evaluated their methods using human study rather than evaluation metrics. 
However, human-oriented assessments widely used in the face restoration field have fatal limitations: \textit{first}, they are unscalable, \textit{second}, a number of assessors and their (large) variances between each other and, \textit{third}, cost of conducting the assessment. 
The absence of appropriate face-oriented metrics results in significant expense and time for evaluation, which is becoming one of the major bottlenecks for the emerging face restoration field.
A question naturally arises in this context is whether we need a face-specific evaluation metric.
% In general, there are numerous object categories in the world, so it is impractical to suggest a metric specific to a particular object category.
Crucially, the face domain is different from conventional object categories in ImageNet~\cite{russakovsky2015imagenet} or COCO~\cite{lin2014microsoft} because of its unique properties (\eg geometry and textures) and a variety of downstream tasks \cite{im2018scale,deng2019arcface,makhmudkhujaev2021re}.
We argue that face images should be evaluated differently from general image domains in terms of image quality assessment (IQA).
The findings of early psychological studies~\cite{Kanwisher4302, Farah1995face, Tsao2008face}, in which human brains use different areas (\ie fusiform face area) to recognize common objects and faces, also support our claim.

This paper introduces a novel face-oriented metric called interpretable face quality assessment (IFQA) based on an adversarial network \cite{goodfellow2014generative}.
The generator, which is a plain face restoration model, attempts to restore high-quality images from low-quality images.
% Rather than simply distinguishing between low and high-quality images, the discriminator is designed to be more sophisticated.
Sub-regions from high-quality images provide `real' supervision to the discriminator, whereas low-quality images and regions from restored face images by the restoration model provide `fake' supervision. 
Inspired by human face perception \cite{Tsao2008face} in which facial primary regions (\eg eyes, nose, and mouth) have a great effect on human face perception, we propose facial primary regions swap (FPRS) that places a greater emphasis on facial primary regions. 
Unlike existing mix-based augmentations~\cite{devries2017improved,Cutmix,CutBlur} that randomly extract local patches from arbitrary positions, FPRS changes regions within a set of the facial primary regions. Additionally, our U-shaped architecture allows us to produce not only single image-level quality scores but also interpretable per-pixel scores.
The proposed metric is related to face image quality assessment (FIQA) \cite{Ortega19}. 
In contrast to FIQA approaches mainly rely on face recognition systems, our IFQA can be considered a more generalized face-oriented metric independent of a specific high-level task.

% Our key idea of giving a lot of weight to the facial primary regions  was not addressed in the existing FIQA, and we show its effectiveness through extensive ablation studies.
We make it clear that our framework aims to realize unresolved face-oriented metrics despite numerous demands raised by existing face restoration studies. 
In our evaluations across various architectures and scenarios, our proposed metric shows higher correlations with human cognition than general IQA metrics and state-of-the-art FIQA metrics.
The contributions of this study are as follows:

\begin{itemize}
    \item
   We propose a new dedicated framework for the face-specific metric that considers the importance of the face primary regions, such as eyes, nose, mouth.
        %\vspace{1.5mm}
    \item Our face-oriented metric matches human judgment significantly more than existing
    general no-reference and state-of-the-art FIQA metrics.
    %\vspace{1.5mm}
        \item Pixel-level evaluation scores enable interpretable image quality analysis that cannot be provided by traditional single-score-based metrics.
    %\vspace{1.5mm}
    % \item We perform human studies with 6,000 human responses and extensively demonstrate the effectiveness of PFQE metric across various experimental settings.
\end{itemize}

%-------------------------------------------------------------------------
\section{Related Work}

\subsection{Face Image Restoration}
% Unlike general image restoration, face image restoration methods, using facial prior knowledge, are specialized for face images.
A series of face image restoration methods have been proposed for addressing certain types of facial image degradation, such as low-resolution, noise, and blur \cite{ZhuECCV16, YuPorikli2016,hong2019unsupervised,Zhang_CopyGAN}.
Although previous studies have shown promising results, they exhibit poor performance in real-world images with unknown and complex degradation.
Some blind face restoration (BFR) approaches have been proposed to address this issue \cite{li2018learning}. 
Prior BFR studies utilized face-specific priors, such as facial component dictionaries \cite{li2020blind}, facial parsing maps \cite{chen2021progressive}, high-quality guided images \cite{li2018learning}.
Among them, GAN inversion approaches based on StyleGAN~\cite{StyleGANv1_FFHQ, karras2020analyzing} have shown promising results~\cite{PULSE, GPEN, GFP-GAN}. Despite significant advances in face restoration methodologies, evaluation metrics, which cannot reflect facial characteristics, are borrowed from general image restoration.
Therefore, state-of-the-art studies have conducted costly human studies to demonstrate the superiority~\cite{PULSE, GPEN, Wan_OldPhoto} of their models, which highly motivates this study.

\subsection{General Image Quality Assessment}
Image quality assessment (IQA) aims to measure the perceptual quality of images and existing approaches can be categorized into two groups: full-reference (FR-IQA) and no-reference (NR-IQA).
FR-IQA evaluates the statistical or perceptual similarity between restored images and reference images. PSNR~\cite{PSNR} and SSIM~\cite{SSIM} are widely used to evaluate face restoration models~\cite{hore2010image}. 
Perceptual metrics have been introduced to alleviate the large semantic gap between traditional FR-IQA and human cognition~\cite{LPIPS, NLPD, DISTS}.
Although the aforementioned metrics are reasonable choices for measuring restoration results, they cannot be applied to real-world scenarios without reference images.

A series of NR-IQA approaches—BRISQUE~\cite{Brisque}, NIQE~\cite{NIQE}, and PI~\cite{srmetricPI}—have been devised to measure the naturalness of images in the blind image quality assessment.
Along with the successful application of NR-IQA in natural scenes, there have been attempts to apply NR-IQA to the BFR problem.
However, numerous BFR studies~\cite{GPEN, Wan_OldPhoto, PULSE} discovered that general NR-IQA metrics have limitations for assessing restored facial images; and therefore, they conduct human studies for evaluation. Consequently, the absence of appropriate evaluation metrics for face restoration requires substantial expenses and becomes a major bottleneck in the face image restoration field. 
To address this critical issue, we propose an evaluation metric specifically designed to focus on facial primary regions. 
% The proposed metric is more consistent with human visual perception than conventional general metrics and provides interpretable visualization.

\subsection{Face Image Quality Assessment}
Face image quality assessment (FIQA) supports face recognition systems by deciding whether or not to discard LQ face images as a preprocessing step.
This process builds stable and reliable face recognition systems in real-world scenarios (\eg surveillance cameras and outdoor scenes).
% Recently, with the rise of face recognition and related fields, various FIQA approaches have been actively proposed~\cite{cosface, deng2019arcface}.
Early FIQA studies exploited analytics-based methods while recent FIQA studies have concentrated their efforts on a learning-based strategy that generates image quality scores directly from face recognition models \cite{Ortega19, SER-FIQ, SDD-FIQA}.
Although previous FIQA studies have shown remarkable results compared with general no-reference IQA \cite{Brisque, NIQE, srmetricPI}, they solely focused on the face recognition task.
% For successful FIQA, intra-class or inter-class information widely used in face recognition should be utilized.
% Thus, face images and the label information that can distinguish the identity of each face image are required to train FIQA models, which limits data selection.
Crucially, FIQA aims to assess the quality of a face image from the point of view of its use in face recognition tasks, involving objective functions derived from face recognition.
Unlike FIQA, our metric does not focus on specific face-related tasks, and thus can be considered a more generalized face image evaluation assessment.
% To the best of our knowledge, PFQE is the first face-centric evaluation metric that focuses on the facial primary regions rather than the entire face area.
% We empirically demonstrate that the proposed metric significantly outperforms all baseline FIQA approaches.

%------------------------------------------------------------------------
\section{Proposed Metric}
\subsection{Pilot Study} 
\begin{figure}
\centering
\small
\begin{tabular}{c@{\hskip 1.5pt}c@{\hskip 1.6pt}c@{\hskip 1.6pt}c@{\hskip 1.6pt}c}
\footnotesize{Reference} & 
\includegraphics[valign=m, width=0.195\linewidth]{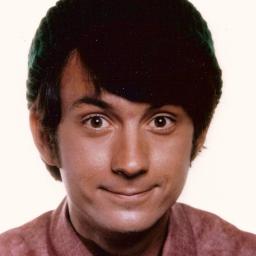}&
\includegraphics[valign=m, width=0.195\linewidth]{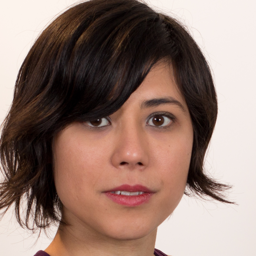}&
\includegraphics[valign=m, width=0.195\linewidth]{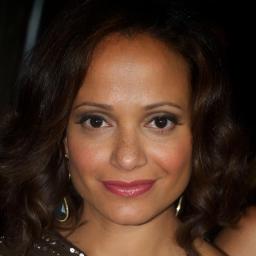}&
\includegraphics[valign=m, width=0.195\linewidth]{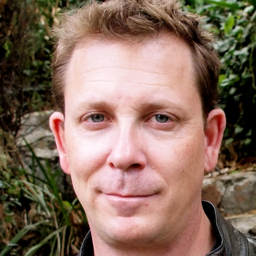}\\
\footnotesize{Image A}&
\includegraphics[valign=m, width=0.195\linewidth]{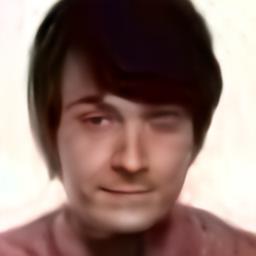}& 
\includegraphics[valign=m, width=0.195\linewidth]{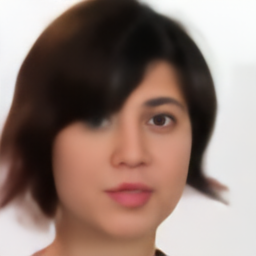}&
\includegraphics[valign=m, width=0.195\linewidth]{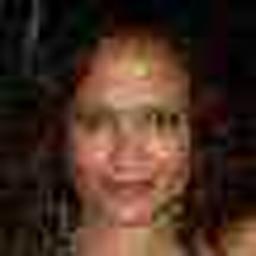}&
\includegraphics[valign=m, width=0.195\linewidth]{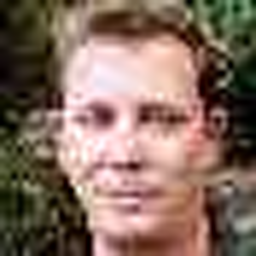}\\
\footnotesize{Image B} &
\includegraphics[valign=m, width=0.195\linewidth]{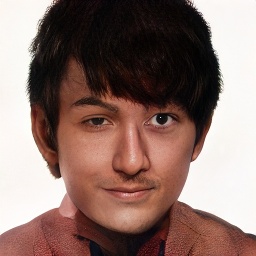}&
\includegraphics[valign=m, width=0.195\linewidth]{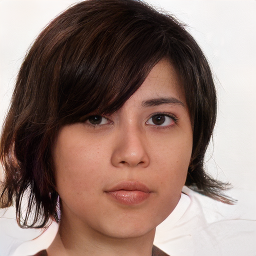}&
\includegraphics[valign=m, width=0.195\linewidth]{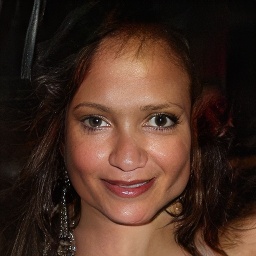}&
\includegraphics[valign=m, width=0.195\linewidth]{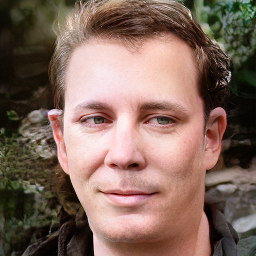}\\\\
\end{tabular} 
    \caption{Comparison of PSNR/SSIM and human assessment on restored face images.
    PSNR/SSIM provides higher scores to `Image A' than `Image B' while human subjects vote `Image B' as higher quality face images than `Image A'.
    }
            % \vspace{-1.5mm}
    \label{fig:preliminary}
\end{figure}

In the preliminary experiments for the assessment of face restoration results, we discovered that facial primary regions play a critical role in human visual perception. 
Most of the participants in the preliminary human study answered that images from `Image B' (third row) in Figure~\ref{fig:preliminary} are more realistic than the ones shown in `Image A' (second row).
However, PSNR and SSIM metrics score `Image A' higher than `Image B'.
Notably, human visual perception is significantly affected by the overall structure and distortions in facial primary regions such as the eyes and nose as shown in the first and second columns.
% Overall, general-purpose metrics have been verified in various tasks; however, their correlation with human visual perception may not be matched in face images.

\begin{figure*}
    \centering
    \includegraphics[width=0.97\linewidth]{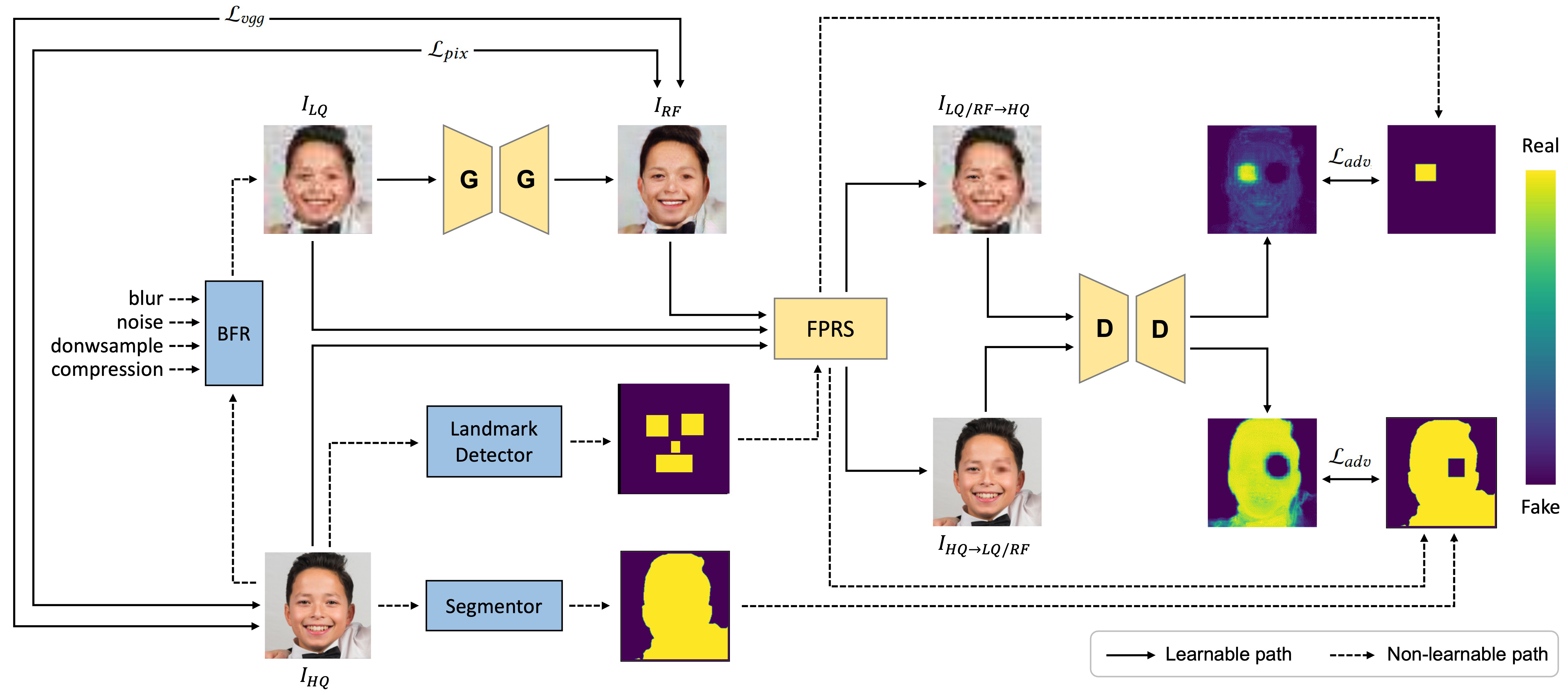}
    \caption{IFQA framework outline.  Given HQ images, we obtain LQ images via BFR formulation. The generator (G) mimics face restoration models, while the discriminator (D) is used to evaluate image quality by determining high-quality regions as `real’ and low-quality or restored regions as `fake'.
    Through its U-Net architecture, the discriminator is able to evaluate the image pixel-by-pixel. 
    FPRS allows the proposed metric to give more weight to facial primary regions that have a significant impact on human visual perception.
    % Once the PFQE framework is trained, we leave only the per-pixel discriminator and use it as a metric function. 
    }
    \label{fig:model_architecture}
\end{figure*}

\subsection{Proposed Framework}
Motivated by the observation in the pilot study, we introduce an evaluation metric considering facial characteristics.
The overall framework is illustrated in Figure~\ref{fig:model_architecture}.

\noindent{\textbf{Generator for image restoration:}} 
The generator consisting of a simple encoder-decoder architecture can be considered a plain face restoration model that outputs restored face images.
The generator is trained to restore LQ images to 256 {$\times$} 256 HQ images. 
During the training phase, we deliberately corrupt HQ images in the FFHQ dataset to make input LQ images as the following BFR formulation:
\begin{equation} %\label{Eq:BFR_LQ}
I_{LQ} = ((I_{HQ} \otimes k)\downarrow_{r} +\ n_{\sigma})_{JPEG_{q}},
 \label{eq:bfr_formulation}
\end{equation}
where ${k}$ is a kernel randomly selected between Gaussian and motion-blur kernel. 
Factors of downsampling, Gaussian noise, and JPEG compression are denoted as ${r}$, ${n_{\sigma}}$ and $q$.
Following previous BFR studies~\cite{li2020blind, GPEN, GFP-GAN}, the range of each factor is  set as $r$: $\left[0.4, 0.9\right)$, $n_{\sigma}$: $\left[50, 250\right)$, $q$: $\left[5, 50\right)$.

\noindent{\textbf{Discriminator for quality assessment:}} 
The discriminator aims to evaluate the quality of query images trained in an adversarial manner with the generator.
We design our discriminator to output per-pixel scores using U-Net-based architecture \cite{UNetGAN} to enhance the generalization ability.
This architecture enables us to classify the regions from HQ images as `real' while the regions from LQ or RF images as `fake'. 
Quality score as a single value (\ie an image-level score) can be obtained via aggregating pixel-level scores. 
Notably, unlike traditional discriminators with adversarial training, we provide `fake' supervision to the generator's input (\ie LQ images) as well as the generator's output (\ie RF images).
We only consider the face region in HQ images as `real' labels instead of entire HQ images. This simple trick helps our metric to focus more on the face region than the entire image. 
In the ablation study, we experimentally show that the `real' supervision setting using only the face region has higher correlations with human judgment than the global region.
We exploit the off-the-shelf face segmentation model~\cite{deeplabv3} pre-trained on CelebAMask-HQ~\cite{CelebAMask-HQ} to obtain binary facial masks from the images.

\begin{figure}
\small
    \centering 
    \begin{tabular}{c|c@{\hskip 1pt}c@{\hskip 1pt}c}
    HQ & \footnotesize{CutMix~\cite{Cutmix}}& \footnotesize{FPRS} &  \footnotesize{FPRS w/ F-Mask} \\
    \includegraphics[width=0.21\linewidth]{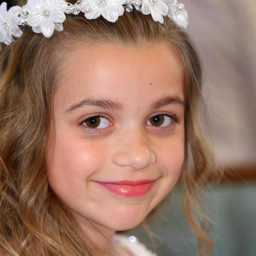}&
    \includegraphics[width=0.21\linewidth]{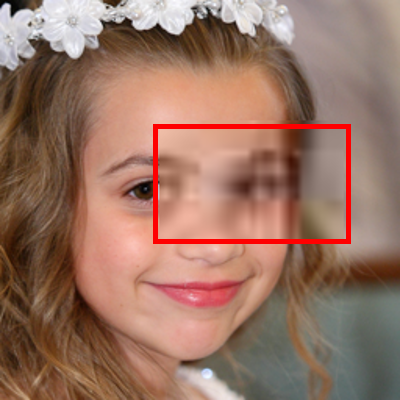}&
    \includegraphics[width=0.21\linewidth]{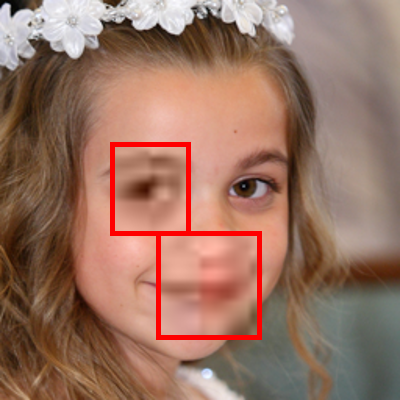}&
    \includegraphics[width=0.21\linewidth]{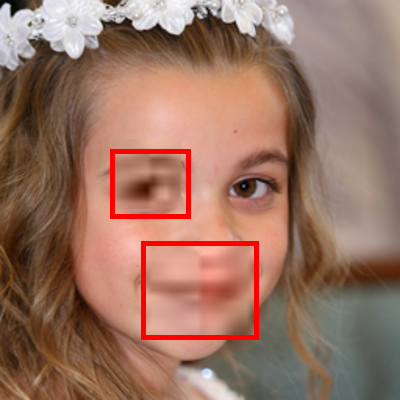}\\
    \includegraphics[width=0.21\linewidth]{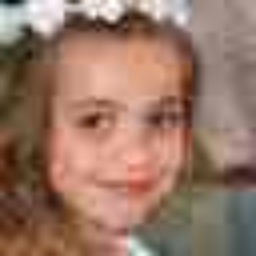}&
    \includegraphics[width=0.21\linewidth]{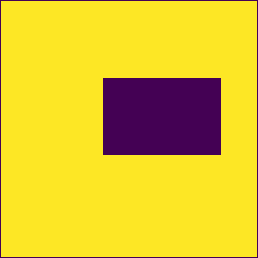}&
\includegraphics[width=0.21\linewidth]{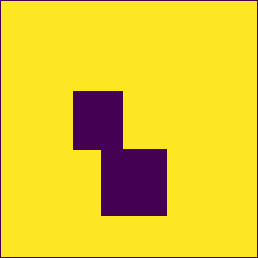}&
    \includegraphics[width=0.21\linewidth]{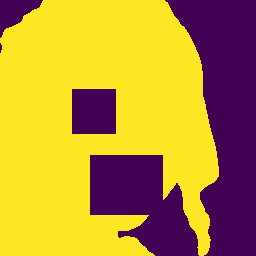}\\
    LQ / RF &  \multicolumn{3}{c}{Pixel-level Supervision}  \\\hline
    \multicolumn{4}{c}{\small{HQ: high-quality \qquad LQ: low-quality \qquad  RF: restored face}} \\
    \end{tabular} \\
    \vspace{1mm}
    \caption{
  Supervision for IFQA metric. 
  Regions from high-quality images provide `real' labels (yellow), while regions from low-quality or restored face images give `fake' labels (purple). 
  The red box indicates the randomly selected swapped region. 
%   Note that our approach utilizes estimated facial region masks and swaps only between facial primary regions, \eg eyes, nose, mouth, via the proposed FPRS.
%   \vspace{-2mm}
  }
    \label{fig:fprs_cutmix}
\end{figure}

Furthermore, we propose a novel augmentation technique called {facial primary regions swap} (FPRS) to reflect facial characteristics to the proposed metric, as shown in Figure~\ref{fig:fprs_cutmix}.  
Firstly, we apply an off-the-shelf landmark detector~\cite{bulat2017far} to HQ images to obtain facial primary regions. Unlike augmentation techniques such as original CutMix for general purpose, we utilize RoIAlign~\cite{He2017Mask} to crop the primary region of facial components. Subsequently, the regions extracted from the LQ or RF images are arbitrarily swapped with the facial primary regions from the HQ images.
Let $I_{LQ/RF}$ denote an LQ or RF image and $I_{HQ}$ denote an HQ image.
Through FPRS operation, we can generate a new image pair to be used for discriminator training as follows: 
\begin{equation} \label{Eq:FPRS}
\begin{aligned}
 {I}_{HQ \rightarrow LQ/RF} &= \mathbf{M_{FPRS}} \odot I_{HQ} \\&+ (\mathbf{1} - \mathbf{M_{FPRS}})   \odot I_{LQ/RF} \\
 {I}_{LQ/RF \rightarrow HQ} &= \mathbf{M_{FPRS}}  \odot I_{LQ/RF} \\&+ (\mathbf{1} - \mathbf{M_{FPRS}}) \odot I_{HQ}, 
\end{aligned}
\end{equation}
where $\mathbf{M_{FPRS}} \in \{0, 1\}^{H\times W}$ is a binary mask for randomly selected facial primary regions. $\mathbf{1}$ is a binary mask filled with ones. $\odot$ indicates element-wise multiplication.

% We can expect the model using FPRS to be sensitive when the input image has an unrealistic shape on the crucial facial regions.

\noindent{\textbf{Objective function:}} 
The IFQA framework is trained by least-square-based adversarial learning~\cite{LSGAN} between the generator and discriminator. The generator is trained to fool the discriminator, and the objective function for the generator is defined as follows:
\begin{equation} \label{Eq:3}
     \mathcal{L}_{adv, G} = \mathbb{E}_{I_{RF}}[(D^{U}(I_{RF}) - \mathbf{1}))^{2}],
\end{equation}
where $D^{U}(\cdot)$ refers to U-Net-based discriminator that outputs per-pixel scores.
Also, we adopt pixel loss to enforce the generator to make $I_{RF}$ to be similar to the corresponding HQ image. The pixel loss compares all of the pixel values between the $I_{RF}$ and HQ images as follows:
\begin{equation} \label{Eq:4}
     \mathcal{L}_{pix} = \mathbb{E}_{I_{RF}, I_{HQ}}[||I_{RF} - I_{HQ}||_{2}].
\end{equation}
To produce photo-realistic facial images, we leverage perceptual loss~\cite{PercepLoss} using the weights of the pre-trained VGG-19 as follows:
\begin{equation} \label{Eq:5}
     \mathcal{L}_{vgg} = \sum_{i}^{}{||f_{i}({I}_{RF})-f_{i}(I_{HQ})||_1},
\end{equation}
where $f_{i}(\cdot)$ is the $i$-th feature extracted from the pre-trained VGG-19 network. Specifically, we use pooling layers in five convolutional blocks for perceptual loss. 

Meanwhile, the objective function of the discriminator is defined as follows:
\begin{equation} \label{Eq:LSGAN_D}
\begin{aligned}
\mathcal{L}_{adv, D} &= \mathbb{E}_{I_{HQ}}[(D^{U}(I_{HQ}) - \mathbf{M_{FACE}})^{2}]\\
&+ \mathbb{E}_{I_{LQ}, I_{RF}}[D^{U}(I_{LQ/RF})^{2}],
\end{aligned}
\end{equation}
where $I_{HQ}$ and $\mathbf{M_{FACE}}$ are HQ images and facial binary masks filled with $1$s only for the face area, respectively.
$I_{LQ/RF}$ is LQ or RF images in which $I_{RF}=G(I_{LQ})$.
The full objective function can be summarized as follows:
\begin{equation} \label{Eq:full}
\begin{aligned}
\underset{G}{\text{min}}~\underset{D}{\text{max}} ~~~ \mathcal{L}_{adv} + {\lambda_{1}}\mathcal{L}_{pix} + {\lambda_{2}}\mathcal{L}_{vgg},
\end{aligned}
\end{equation}
where $\mathcal{L}_{adv} = \mathcal{L}_{adv, G} + \mathcal{L}_{adv, D}$, and $\lambda_{1}$ and $\lambda_{2}$ are scaling parameters. We set $\lambda_{1}$ and $\lambda_{2}$ as 50 and 5, respectively.

\noindent{\textbf{Assessment protocol:}} 
Once the IFQA framework is trained, we only use the per-pixel discriminator for image quality assessment.
The pixel-level assessment score of the discriminator enables us to perform an interpretable in-depth analysis.
Given an input image $I$, we can obtain an image-level quality score ($QS$) by simply averaging every pixel-level score from the per-pixel discriminator as:
\begin{equation} \label{Eq:eval}
      QS = \frac{1}{H\times W}\sum_{i=1}^{H}{\sum_{j=1}^{W}{D^{U}_{i,j}(I)}}
\end{equation}
\section{Experiments}
\subsection{Implementation Details}
\noindent{\textbf{Datasets:}} 
% The PFQE metric intrinsically should not be highly dependent on a specific dataset. 
% We separate the training dataset for the learning-based metric and the test dataset to prove the generalization of our metric.
Randomly selected 20,000 images from FFHQ~\cite{StyleGANv1_FFHQ} were used for training the IFQA framework. 
Meanwhile, we constructed three types of benchmark test datasets.
First, we construct a test set by combining CelebA-HQ~\cite{CelebAHQ} and FFHQ images with high-quality images and considerable variations, widely used in face restoration tasks.
% Based on the data split, the images used for training and those used for testing do not overlap in the FFHQ dataset.
Given high-quality images of the test set, we obtained low-resolution query images using the BFR formulation Eq.~\ref{eq:bfr_formulation}.
Second, considering real-world scenarios, we constructed a test set using in the wild face (IWF)~\cite{GPEN}, which is widely used for BFR problems.
Notably, the IWF dataset only provides low-quality images without high-quality reference images.
Third, CelebA-HQ, FFHQ, and IWF were combined and used as a test set for the ablation study to demonstrate the generalization ability of the proposed metric.

% \noindent{\textbf{Framework details:}}
% The generator consists of five residual blocks in the encoder and decoder with skip connections. 
% The discriminator has a similar architecture as the generator. 
% The Adam optimizer was used to train both the generator and discriminator with learning rate $lr = 0.0002$, $\beta_1 = 0.5$, $\beta_2 = 0.999$. The mini-batch size was set to 20. 
% More details can be found in the supplementary material.

\noindent{\textbf{Image restoration models:}} 
We used various image restoration models including general image restoration models (\eg RCAN~\cite{RCAN}, DBPN~\cite{DBPN}) and face restoration models (\eg HiFaceGAN~\cite{Yang_HiFaceGAN}, DFDNet~\cite{li2020blind}, GPEN~\cite{GPEN}) to evaluate the proposed metric quantitatively and qualitatively.
% For a fair comparison, we used each model's official training code and hyperparameters. 
% Because DFDNet did not release the training code, we trained DFDNet using the same training protocol as much as possible based on the descriptions of the paper. 
% Additionally, we used the state-of-the-art GPEN model pre-trained on FFHQ.

% \noindent{\textbf{Comparative IQA metrics:}} 
% We used a series of assessment metrics commonly used in face restoration tasks: 
% three traditional FR-IQA metrics (\eg PSNR~\cite{PSNR}, SSIM~\cite{SSIM}, MS-SSIM~\cite{MS-SSIM}), perceptual FR-IQA metric such as LPIPS~\cite{LPIPS}), two conventional NR-IQA metrics (\eg BRISQUE~\cite{Brisque}, NIQE~\cite{NIQE}), and perceptual NR-IQA metric such as PI~\cite{srmetricPI}). We compared with four state-of-the-arts FIQA methods (\eg FaceQnet-V0~\cite{Ortega19}, FaceQnet-V1~\cite{faceqnetv1}, SER-FIQ~\cite{SER-FIQ}, and SDD-FIQA~\cite{SDD-FIQA}). All the compared methods were reproduced from their official codes, and we describe the details of experimental settings in the supplementary material.

\subsection{Quantitative Analysis}
\noindent{\textbf{Human study protocol:}} 
For quantitative comparison with the proposed IFQA metric and the existing IQA metrics, we conducted a human study of ranking the realistic facial images from given images.
We carefully designed survey questions and prepared face images to use to estimate human visual judgments. 
One sample consisted of six images, including an LQ image and one image each restored from RCAN, DBPN, DFDNet, HiFaceGAN, and GPEN.
All 200 sample images were randomly selected from the FFHQ, CelebA-HQ, and IWF datasets. 
The total number of images for the human study was 1,200.

\begin{figure}
    \centering
    \includegraphics[width=1\linewidth]{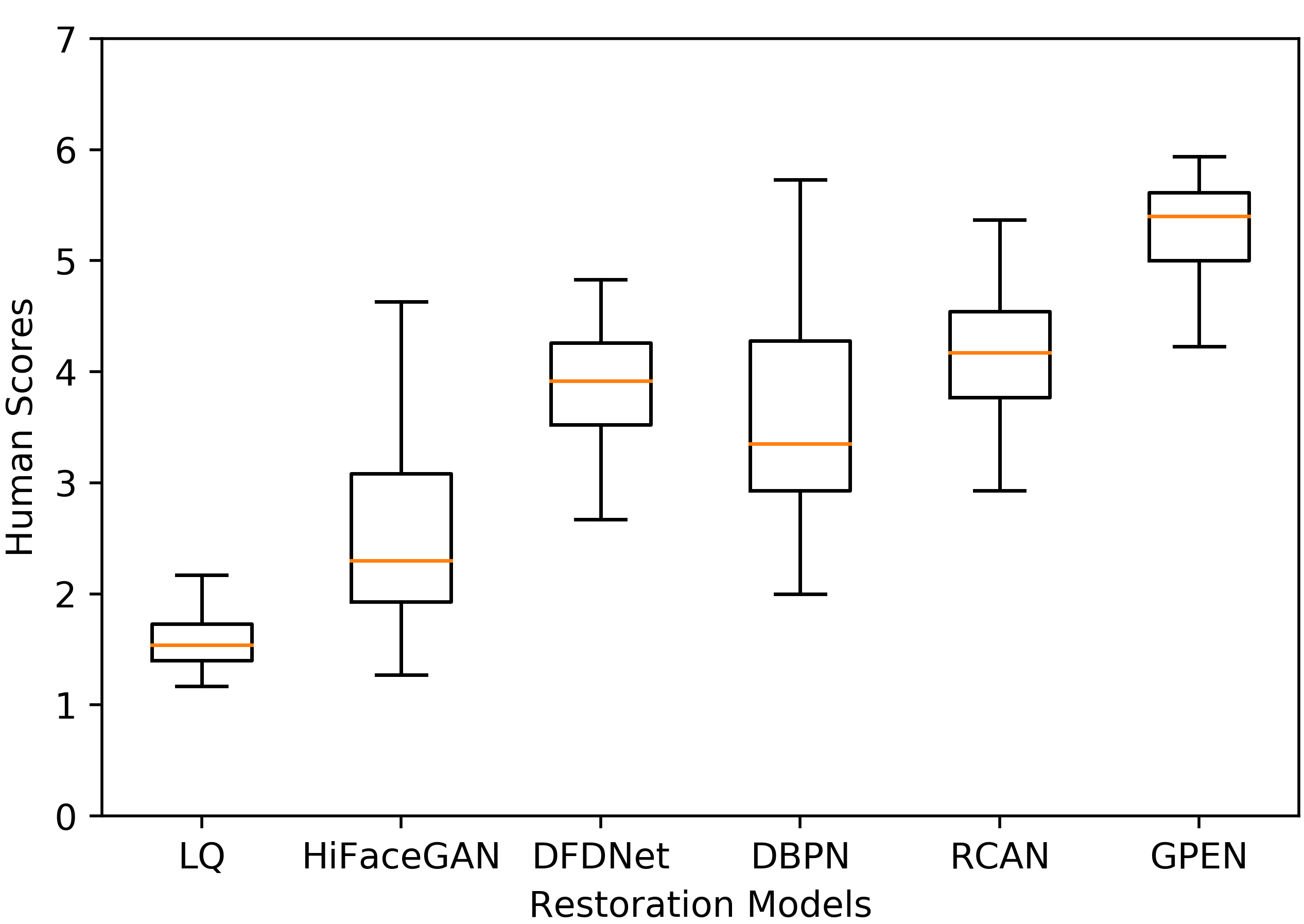}
    \caption{Box plot of restoration models through human study.}
    \label{fig:box_plot}
\end{figure}

We asked participants to rank given samples from closest to furthest to a realistic human face. 
We used Amazon Mechanical Turk crowdsourcing \cite{paolacci2010running} to gather participant's responses systematically.
We also received responses from researchers who are majors in various AI-related fields and are not directly related to this study.
We assigned 30 subjects per sample, and the total number of responses across all samples was 6,000.
The final rank of each sample was calculated by the weighted average rank. 
 Figure~\ref{fig:box_plot} present the box plot of our human study results.
% We describe additional details in the supplementary material.

\begin{figure*}[t]
    \centering
    \begin{tabular}{c@{\hskip 1pt}c@{\hskip 1pt}c@{\hskip 1pt}c@{\hskip 1pt}c@{\hskip 1pt}c@{\hskip 1pt}c@{\hskip 1pt}c@{\hskip 1pt}c}
    \small{LQ\textsubscript{Blur}}& \small{LQ\textsubscript{Down$\downarrow$}}& \small{LQ\textsubscript{Mix}}& \small{HiFaceGAN}& \small{DFDNet}& \small{DBPN}& \small{RCAN}& \small{GPEN}& \small{Reference}\\
\includegraphics[width=0.105\linewidth]{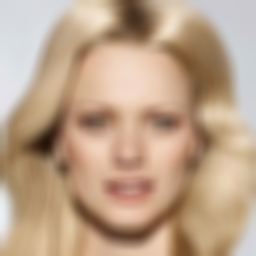}&
\includegraphics[width=0.105\linewidth]{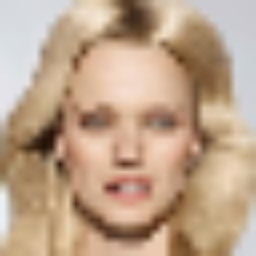}&
\includegraphics[width=0.105\linewidth]{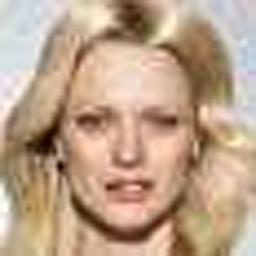}&
\includegraphics[width=0.105\linewidth]{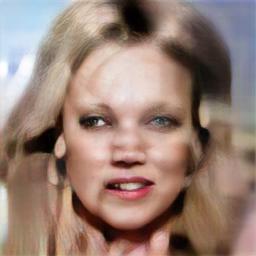}&
\includegraphics[width=0.105\linewidth]{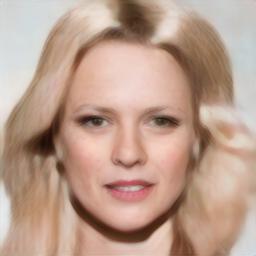}&
\includegraphics[width=0.105\linewidth]{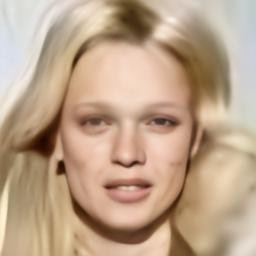}&
\includegraphics[width=0.105\linewidth]{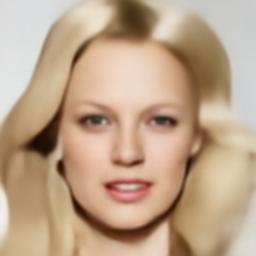}&
\includegraphics[width=0.105\linewidth]{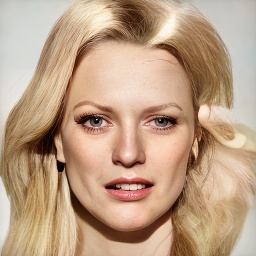}&
\includegraphics[width=0.105\linewidth]{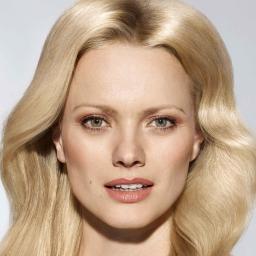}\\
\includegraphics[width=0.105\linewidth]{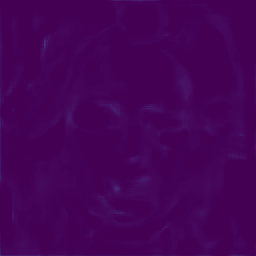}&
\includegraphics[width=0.105\linewidth]{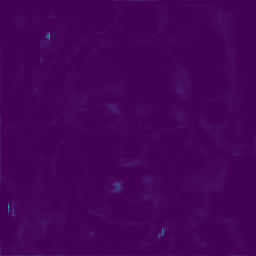}&
\includegraphics[width=0.105\linewidth]{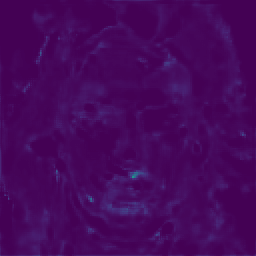}&
\includegraphics[width=0.105\linewidth]{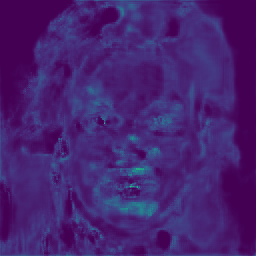}&
\includegraphics[width=0.105\linewidth]{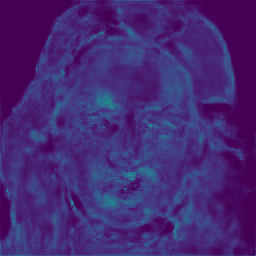}&
\includegraphics[width=0.105\linewidth]{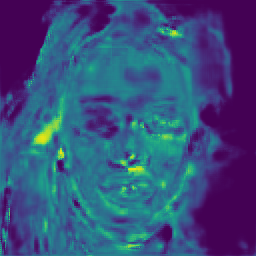}&
\includegraphics[width=0.105\linewidth]{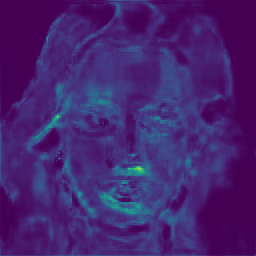}&
\includegraphics[width=0.105\linewidth]{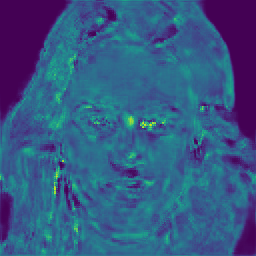}&
\includegraphics[width=0.105\linewidth]{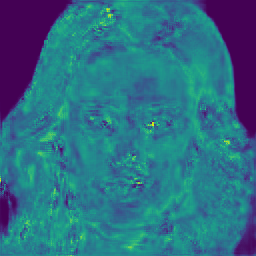}\\
\includegraphics[width=0.105\linewidth]{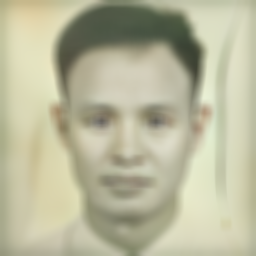}&
\includegraphics[width=0.105\linewidth]{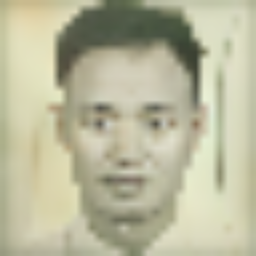}&
\includegraphics[width=0.105\linewidth]{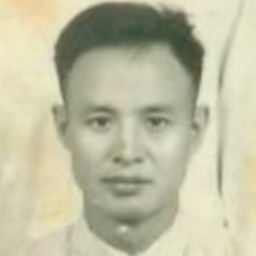}&
\includegraphics[width=0.105\linewidth]{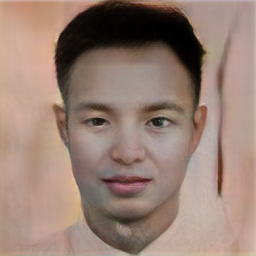}&
\includegraphics[width=0.105\linewidth]{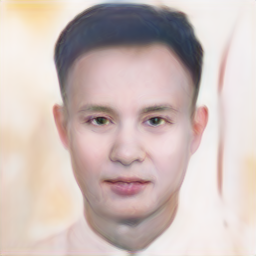}&
\includegraphics[width=0.105\linewidth]{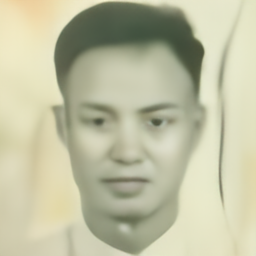}&
\includegraphics[width=0.105\linewidth]{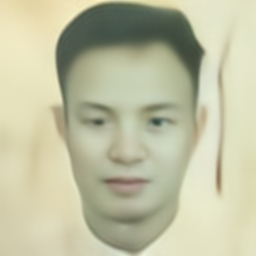}&
\includegraphics[width=0.105\linewidth]{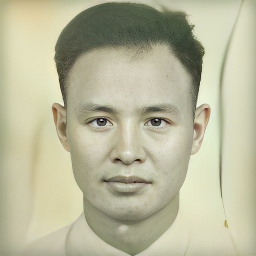}&
\includegraphics[width=0.105\linewidth]{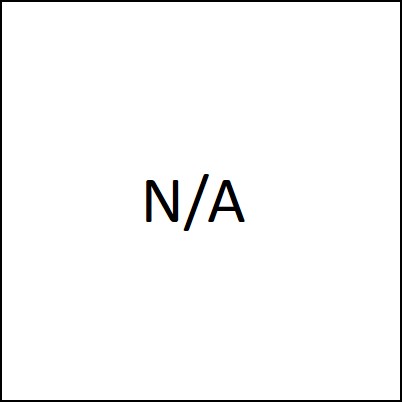}\\
\includegraphics[width=0.105\linewidth]{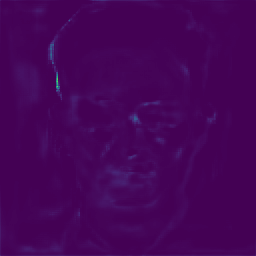}&
\includegraphics[width=0.105\linewidth]{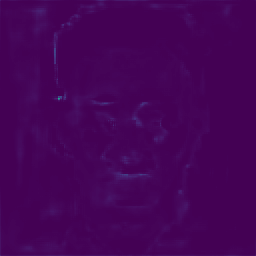}&
\includegraphics[width=0.105\linewidth]{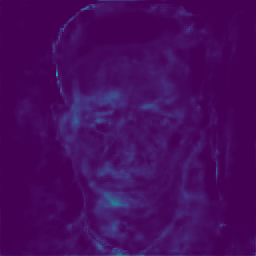}&
\includegraphics[width=0.105\linewidth]{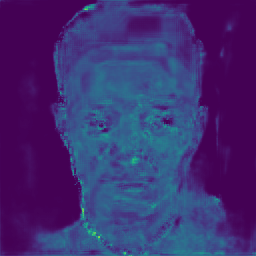}&
\includegraphics[width=0.105\linewidth]{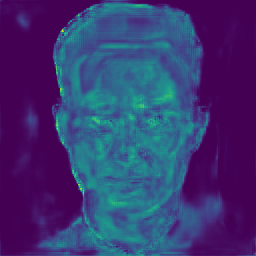}&
\includegraphics[width=0.105\linewidth]{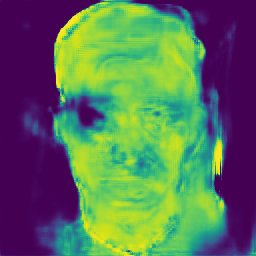}&
\includegraphics[width=0.105\linewidth]{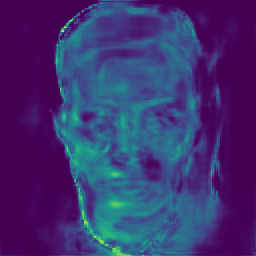}&
\includegraphics[width=0.105\linewidth]{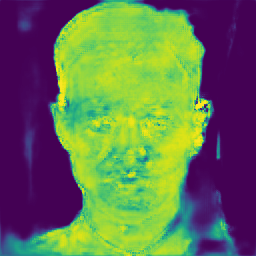}&
\includegraphics[width=0.105\linewidth]{Figure_Qualitative/N_A_Big.jpg}\\
    \end{tabular}
    \caption{
    Interpretable visualization of the proposed metric on various types of LQ images, HQ images (\ie, reference), and RF images from the restoration models.
    The first and second rows show images from FFHQ and their corresponding interpretability maps, respectively.
    The third and fourth rows present pairs from IWF that does not provide reference images. Brighter area indicates the higher quality.}
            % \vspace{-1.5mm}
    \label{fig:qualitative}
\end{figure*}

\noindent{\textbf{IQA comparative analysis:}}
Once human ranking responses were obtained for each sample, we measured Spearman's rank order correlation coefficient (SRCC)~\cite{SRCC} and Kendall rank order correlation coefficients (KRCC)~\cite{KRCC} for quantitative analysis.
These approaches are widely used to measure the correlation between human judgments and other metrics across various fields.

\noindent{\textbf{Aanalysis on FFHQ \& CelebA-HQ:}}
First, we performed a comparative evaluation on FFHQ and CelebA-HQ with the existing no-reference IQA (NR-IQA) metrics, including state-of-the-art FIQA metrics.
Table~\ref{tab:SRCC_KRCC_Comparison_FRIQA} shows that IFQA has the highest correlations with human preferences in both SRCC and KRCC metrics.
Interestingly, NIQE, the widely used NR-IQA metric, shows the lowest correlation.
We compared with the recent FIQA methods, which are specially designed for face recognition.
IFQA is superior to other FIQA metrics. 
Evidently, our face-oriented metric is steadily more consistent with human judgment compared with the existing NR-IQA metrics.

Although the proposed IFQA is an NR-IQA metric, we conducted a comparative analysis with full-reference metrics (\ie FR-IQA) to prove the general applicability of the proposed metric.
The traditional but widely used metrics, PSNR~\cite{PSNR} and SSIM~\cite{SSIM}, show values less than 0.2 in both SRCC and KRCC.
% showing weak MS-SSIM\cite{MS-SSIM} shows a value of 0.3571 for SRCC and 0.2799 for KRCC, which is relatively better than the traditional FR-IQA metric, but still shows a low correlation compared to ours.
Perceptual metric, LPIPS~\cite{LPIPS}, shows 0.6685 for SRCC and 0.5560 for KRCC, showing better performance than the existing FR-IQA metrics.
Crucially, all FR-IQA metrics cannot be applied to practical scenarios in the wild in which there is no reference image.
The proposed IFQA shows the highest correlations with the human perception among NR-IQA metrics and is comparable to the state-of-the-art FR-IQA metric despite not requiring any reference image.

\begin{table}[]
\small
    \centering
    \caption{Comparative analysis on FFHQ and CelebA-HQ. 
    }
    \begin{tabular}{l|c|c|c}\hline
    \toprule
     Metric& Type & SRCC $\uparrow$& KRCC $\uparrow$\\\hline\hline
     NIQE~\cite{NIQE}& \multirow{4}{*}{\makecell{NR-IQA \\ (General)}} &0.2668& 0.2039 \\ %\hline 
     PI~\cite{srmetricPI}& &0.4125& 0.3173 \\
     BRISQUE~\cite{Brisque}&  &0.4405 & 0.3373 \\
     IFQA \textbf{(Ours)} & & \textbf{0.6400}& \textbf{0.5186}\\ \hline 
     SER-FIQ~\cite{SER-FIQ}& \multirow{4}{*}{NR-IQA (FIQA)} &0.3554 & 0.2706 \\
     FaceQnet-V1~\cite{faceqnetv1}& &0.4560& 0.3453\\
     FaceQnet-V0~\cite{Ortega19}&  &0.5491& 0.434 \\
     SDD-FIQA~\cite{SDD-FIQA}& &0.5920 & 0.4840\\\hline
    %  PSNR~\cite{PSNR} &\multirow{4}{*}{FR-IQA}&0.1011 &0.0893\\
    %  SSIM~\cite{SSIM}& &  0.1885& 0.1999\\
    %  MS-SSIM~\cite{MS-SSIM}& & 0.3571& 0.2799\\
    %  LPIPS~\cite{LPIPS} & & \textbf{0.6685} & \textbf{0.5560}\\\hline
    \bottomrule
    \end{tabular}
            % \vspace{-1.5mm}
    \label{tab:SRCC_KRCC_Comparison_FRIQA}
\end{table}

\begin{table}[]
\small
    \centering
    \caption{Comparative analysis on IWF.}
    \begin{tabular}{l|c|c|c}\hline
    \toprule
     Metric& Type & SRCC $\uparrow$& KRCC $\uparrow$\\\hline
     NIQE~\cite{NIQE}&\multirow{4}{*}{\makecell{NR-IQA \\ (General)}} &0.5005& 0.4053\\ 
     PI~\cite{srmetricPI}& &0.6382& 0.5320 \\
     BRISQUE~\cite{Brisque}& &0.6451 & 0.5573 \\
     IFQA \textbf{(Ours)} & &\textbf{0.6988}& \textbf{0.6013}\\\hline
     SER-FIQ~\cite{SER-FIQ}&  \multirow{4}{*}{NR-IQA (FIQA)} &0.1657 & 0.1386 \\ 
     FaceQnet-V1~\cite{faceqnetv1}& &0.2725& 0.2106 \\
     FaceQnet-V0~\cite{Ortega19}& &0.4474& 0.3813\\
     SDD-FIQA~\cite{SDD-FIQA}& &0.5131 & 0.4120 \\
     \hline
    %  PSNR~\cite{PSNR} &\multirow{4}{*}{FR-IQA}&0.1011 &0.0893\\
    %  SSIM~\cite{SSIM}& &  0.1885& 0.1999\\
    %  MS-SSIM~\cite{MS-SSIM}& & 0.3571& 0.2799\\
    %  LPIPS~\cite{LPIPS} & & \textbf{0.6685} & \textbf{0.5560}\\
    \bottomrule
    \end{tabular}
            % \vspace{-1.5mm}
    \label{tab:SRCC_KRCC_Comparison_NRIQA}
\end{table}

\noindent{\textbf{Analysis on a real-world dataset:}}
We measure the correlation value using real-world face images~\cite{GPEN} to prove the generalization ability of the assessment.
We exclude the FR-IQA metrics and use NR-IQA and FIQA metrics for comparison because real-world face images have no HQ reference images.
Table~\ref{tab:SRCC_KRCC_Comparison_NRIQA} shows that existing FIQA metrics show low correlation values except for the recently proposed state-of-the-art SDD-FIQA.
SDD-FIQA metric shows a reasonable correlation value; general NR-IQA metrics have similar or higher values than FIQA metrics.
Our proposed metric shows the highest correlation with human visual perception than general NR-IQA and FIQA metrics on the real-world dataset.
% The results demonstrate the limitations of utilizing FIQA methods in the face image restoration task.

%   \vspace{-1mm}
\subsection{Qualitative Analysis}
\noindent{\textbf{Interpretability evaluation:}}
% One of the main characteristics of PFQE is that it can be visualized by providing interpretable pixel-level scores.
To prove the effectiveness and utilization of the interpretability of the proposed metric, we generated a variety of LQ images (\eg blur, downsampling, and mix) from the HQ reference images based on the BFR protocol~\cite{GPEN, GFP-GAN}.
Subsequently, we restored images by applying widely used general image restoration models (\eg RCAN, DBPN) and face restoration models (\eg DFDNet, HiFaceGAN, GPEN) to the LQ\textsubscript{Mix} image, which is a combination of various degradation factors.

The interpretability maps of IFQA are shown in Figure~\ref{fig:qualitative}.
HiFaceGAN produces plausible eyes, mouth, and teeth arrangements, but it contains several artifacts in the no-reference IWF dataset. 
DFDNet generates reasonable facial structures but fails to restore the details compared with other recent face restoration models. 
DBPN usually incurs an unnatural shape of the facial primary regions or facial contour, causing IFQA to attain low scores in those regions. 
RCAN results in over-smoothed restored images, which leads to an overall lower score than face restoration model results.
GPEN generates the most realistic facial details compared with other models, which results in high scores in facial regions.
We can confirm that the overall results are consistent with the human judgment in Figure~\ref{fig:box_plot}.
Even though our metric shows the highest consistency with human judgment compared to other metrics, there are still limitations.
Since our metric is learning-based on a synthetic dataset, inconsistent results could be produced on real-world data. Incorporating real-world performance degradation into the learning process remains as a future work.

\begin{figure}
\small
    \centering
    \begin{tabular}{cc@{\hskip 2pt}c@{\hskip 2pt}c}
    Reference& Image A& Image B& Image C\\\toprule
    \includegraphics[width=0.21\linewidth]{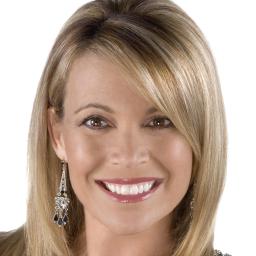}&
    \includegraphics[width=0.21\linewidth]{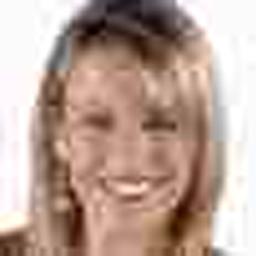}&
    \includegraphics[width=0.21\linewidth]{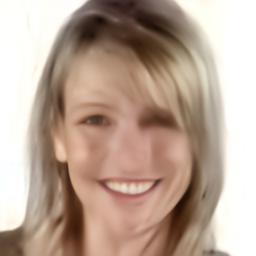}&
    \includegraphics[width=0.21\linewidth]{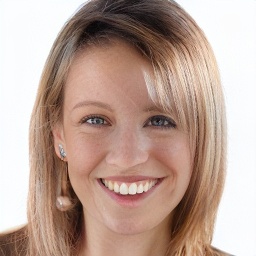}\\
    {PSNR}&
    \includegraphics[valign=m, width=0.21\linewidth]{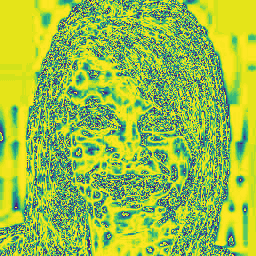}&
    \includegraphics[valign=m, width=0.21\linewidth]{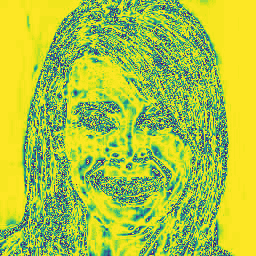}&
    \includegraphics[valign=m, width=0.21\linewidth]{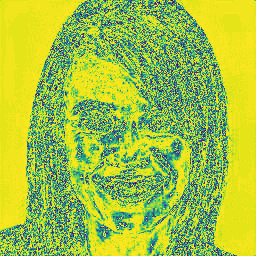}\\
    {SSIM}&
    \includegraphics[valign=m, width=0.21\linewidth]{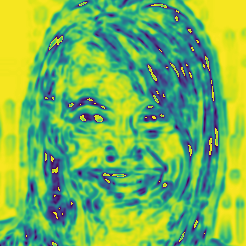}&
    \includegraphics[valign=m, width=0.21\linewidth]{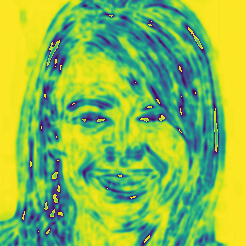}&
    \includegraphics[valign=m, width=0.21\linewidth]{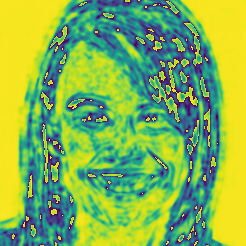}\\
    {Ours}&
    \includegraphics[valign=m, width=0.21\linewidth]{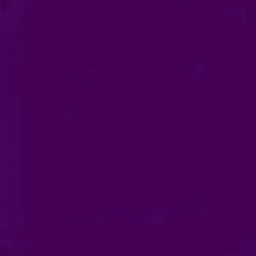}&
    \includegraphics[valign=m, width=0.21\linewidth]{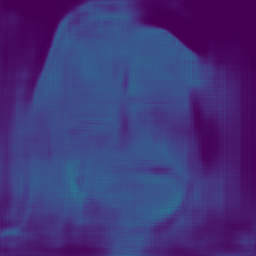}&
    \includegraphics[valign=m, width=0.21\linewidth]{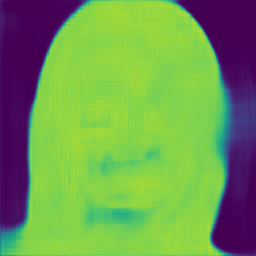}\\
    \end{tabular} 
        \vspace{1mm}
    \caption{Comparison of the proposed metric with PSNR/SSIM with respect to pixel-level score. Bright areas indicate higher similarity to the reference image.
    }
            % \vspace{-1.5mm}
    \label{fig:interpretable_vsPSNR}
\end{figure}

\noindent{\textbf{Comparison with PSNR/SSIM:}} 
The proposed face-oriented metric enables pixel-level visualization, whereas general NR-IQA metrics cannot provide pixel-level scores. 
Traditional FR-IQA metrics, such as PSNR and SSIM, can provide pixel-level scores; however, the results are not interpretable.
We compare PSNR and SSIM maps with the proposed IFQA in terms of pixel-level scores in Figure~\ref{fig:interpretable_vsPSNR}. 
The map of PSNR is obtained by $L2$ distance between the reference image and the restored images. 
For a clear qualitative comparison with IFQA, we reversed the distance map of PSNR and SSIM.
The brighter the area, the more similar it is to the reference image.
In the figure, for the severely degraded `Image A', IFQA attains an overall low score, whereas PSNR and SSIM attain a sparse low score.
For `Image B', which is of low quality except for a tiny part of the face, IFQA scores high in these undamaged regions.
 
\subsection{In-depth Analysis}
% Several design choices exist for the PFQE framework. We conducted an in-depth analysis on FFHQ, CelebA-HQ, and IWF datasets.  
% In Table~\ref{tab:Abl_SRCC_KRCC},~\ref{tab:Abl_generator}, and ~\ref{tab:Abl_discriminator}, we report the averaged values from the three datasets.

\begin{table}[]
\small
    \centering 
\caption{Ablation study of IFQA framework on FFHQ, CelebA-HQ, and IWF datasets. We report the average correlation for the entire test datasets. 
% The same trainable plain face restoration model  is used as the generator across all comparative discriminators.
}
    \begin{tabular}{l|c|c}\hline
    \toprule
    Discriminator & SRCC $\uparrow$& KRCC $\uparrow$\\\hline
    Baseline (single-output)& 0.5885& 0.4840\\
Baseline (per-pixel) & 0.4674& 0.3820\\\hline
%$D$ (Encoder)& 0.5& 0.4133\\
Baseline (per-pixel) + CutMix~\cite{Cutmix}& 0.5437& 0.4420\\ 
Baseline (per-pixel) + FPRS & 0.6265& 0.5213\\
%Baseline + $M_{face}$& 0.6308& \textbf{0.5333}\\
Baseline (per-pixel)  + FPRS + F-Mask &  \textbf{0.6694}& \textbf{0.5600}\\ 
    \bottomrule
         \multicolumn{3}{l}{F-Mask: facial masks using a segmentation model}  \\
    \end{tabular} 
                % \vspace{-1.5mm}
    \label{tab:Abl_SRCC_KRCC}
\end{table}

\noindent{\textbf{Ablation study for main modules:}} 
We present an ablation study considering the following variants: 
(i) a baseline model consisting of a learnable generator and an encoder-based discriminator that outputs a single value,
(ii) a model that differs only in the discriminator from the first baseline, which outputs the pixel-level score,
(iii) a model with original CutMix added to the second baseline model,
(iv) a model with the proposed FPRS added to the second baseline model, and
(v) a model with the facial mask information added to the fourth baseline model (\ie our final model).

Table~\ref{tab:Abl_SRCC_KRCC} reports the quantitative results on FFHQ, CelebA-HQ, and IWF datasets. 
From the table, we can see the following observations.
Without the proposed modules, a single-output discriminator composed of only an encoder is more similar to human judgment than a per-pixel discriminator.
Although CutMix results in performance improvement, it is significantly inconsistent with human judgment compared to our IFQA model with FPRS.
Moreover, adding face mask information to the baseline model results in a metric that is even closer to human perception.

\noindent{\textbf{Generator change analysis:}} 
% Our framework consists of two sub-networks where one is a generator used as a face restoration model while the other one is a discriminator for assessment.
We hypothesize that a trainable naive model as a generator is more suitable for learning the discriminator than pre-trained general or face restoration models.
To prove this hypothesis, we compare our plain model with the four conventional approaches.
% (i) GPEN, a state-of-the-art face image restoration model that uses generative prior,
% (ii) DFDNet, a face image restoration model based on facial components dictionaries,
% (iii) DBPN, a general image restoration model with backprojection mechanism, and
% (iv) RCAN,  a general image restoration model with residual channel-attention networks. 
In Table~\ref{tab:Abl_generator}, our plain U-Net-based generator shows the highest correlation with human perception, whereas cutting-edge GPEN shows the lowest value.

\noindent{\textbf{Discriminator backbone analysis:}} 
Because the proposed IFQA metric depends on the trainable discriminator, we performed a comparison considering the following backbone architectures of the discriminator. 
% U-Net variant using spatial feature transform (SFT)~\cite{sftCVPR18},
% (iii) U-Net variant based on ResNet-50~\cite{HeCVPR16} encoder,
% (iv) U-Net variant based on VGG-16~\cite{VGG16} encoder, and
% (v)  U-Net variant based on VGG-19~\cite{VGG16} encoder.
Table~\ref{tab:Abl_discriminator} shows that all the variants of our metric are superior to most of the existing IQA metrics and our VGG-19-based model produces the best performance.

\begin{table}[]
\small
    \centering
\caption{Performance comparison with respect to generator models on FFHQ, CelebA-HQ, and IWF.}
    \begin{tabular}{l|c|c|c|c}\hline
    \toprule
 Generator  & Task & Parameters  & SRCC $\uparrow$& KRCC $\uparrow$\\\hline
%GPEN~\cite{GPEN}& \multirow{4}{*}{\ding{55}}& 0.4997& 0.4166\\
 GPEN~\cite{GPEN} & FIR &\multirow{4}{*}{pre-trained} &  0.4997& 0.4166\\
 DFDNet~\cite{li2020blind} &FIR &  &0.5391& 0.4366\\
 DBPN~\cite{DBPN} &GIR & &0.5582& 0.4586\\
 RCAN~\cite{RCAN}&GIR & &0.5711& 0.4680\\\hline
 RCAN & GIR & learnable & 0.6454 & 0.5480\\
 Plain model & FIR  &learnable & \textbf{0.6694}& \textbf{0.5600}\\
 
    \bottomrule
      \multicolumn{5}{c}{FIR: face image restoration \quad GIR: general image restoration}  \\
    \end{tabular} 
    \label{tab:Abl_generator}
\end{table}

\begin{table}[]
\small
    \centering
\caption{Performance comparison  with respect to the backbone of discriminators on FFHQ, CelebA-HQ, and IWF.}
    \begin{tabular}{ l | c | c | c }\hline
    \toprule
    Discriminator & Parameters & SRCC $\uparrow$& KRCC $\uparrow$\\
    \hline
U-Net~\cite{UNet15}& \multirow{5}{*}{learnable} & 0.6100 & 0.5006\\
U-Net + SFT~\cite{sftCVPR18}& &  0.6314& 0.5253\\
ResNet-50~\cite{HeCVPR16}&&  0.6311& 0.5346\\ 
VGG-16~\cite{VGG16}& &  0.6365& 0.5286\\
VGG-19~\cite{VGG16}& &  \textbf{0.6694} & \textbf{0.5600}\\ 
    \bottomrule
    \end{tabular} 
    \label{tab:Abl_discriminator}
\end{table}

\subsection{Further Use Case Analysis}

\begin{figure}[t]
\small
    \centering
    \begin{tabular}{c@{\hskip 1pt}c@{\hskip 1pt}c@{\hskip 1pt}c}
    Reference& Image A& Image B& Image C\\
    \includegraphics[width=0.24\linewidth]{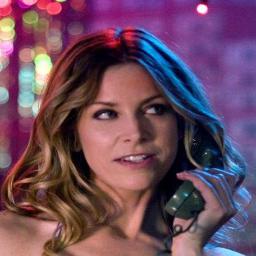}&
    \includegraphics[width=0.24\linewidth]{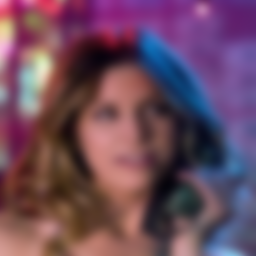}&
    \includegraphics[width=0.24\linewidth]{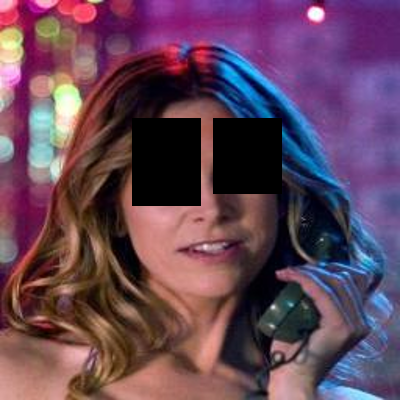}&
    \includegraphics[width=0.24\linewidth]{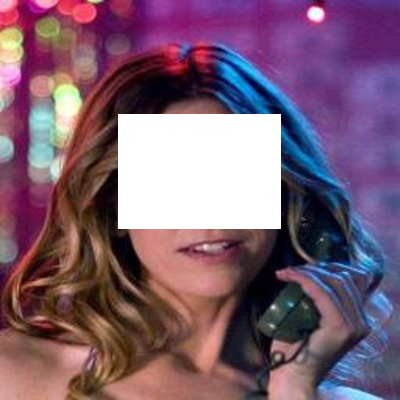}\\
    \includegraphics[width=0.24\linewidth]{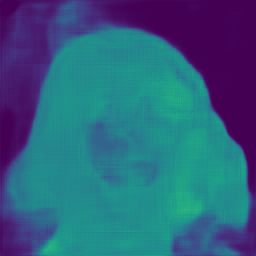}&
    \includegraphics[width=0.24\linewidth]{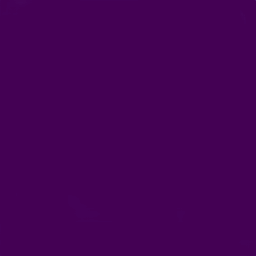}&
    \includegraphics[width=0.24\linewidth]{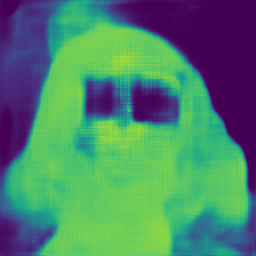}&
    \includegraphics[width=0.24\linewidth]{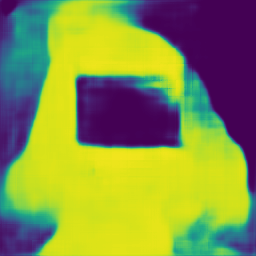}\\
    \end{tabular} 
    \vspace{1mm}
    \caption{IFQA results in more challenging scenarios. Brighter area indicates the higher quality.}
        \vspace{-1mm}
    \label{fig:challenging}
\end{figure}

\noindent{\textbf{IFQA under realistic conditions:}} 
% The proposed metrics addresses commonly used face image datasets under relatively stable conditions such as cropped frontal faces.
% However, real-world face images could include various conditions such as lighting conditions, different viewpoints, different ages, accessories, etc.
We applied various degradations to the images from VGGFace2~\cite{vggface2} to validate the generalization ability of the proposed metric in more challenging scenarios.
In Figure~\ref{fig:challenging}, IFQA provides acceptable results for most scenarios.
Unlike the high-quality reference image, low-quality `Image A' shows low pixel-level scores in entire regions.
Even Black or white box occlusion as seen in (`Image B') and (`Image C') is not considered as facial degradation factors, but interestingly, the proposed metric shows reasonable results.
%Our metric shows low scores for blur in a specific part of the face (`Image B') or distortion across the face and upper body (`Image C').

\begin{figure}[t]
\small
    \centering
    \begin{tabular}{c@{\hskip 1pt}c@{\hskip 1pt}c@{\hskip 1pt}c}
    Reference& Image A& Image B& Image C\\
    \includegraphics[width=0.24\linewidth]{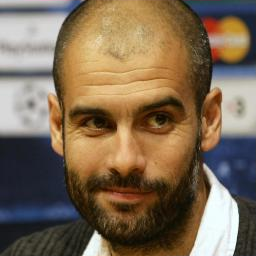}&
    \includegraphics[width=0.24\linewidth]{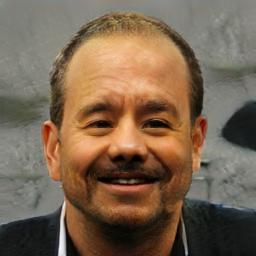}&
\includegraphics[width=0.24\linewidth]{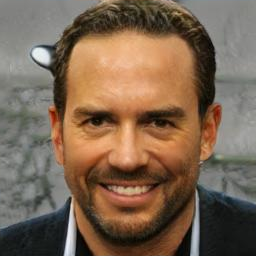}&
    \includegraphics[width=0.24\linewidth]{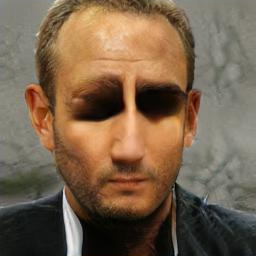}\\
    \includegraphics[width=0.24\linewidth]{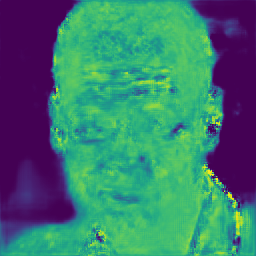}&
    \includegraphics[width=0.24\linewidth]{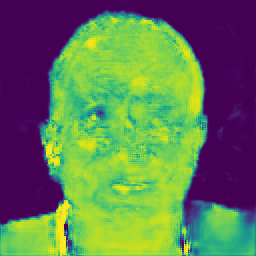}&
    \includegraphics[width=0.24\linewidth]{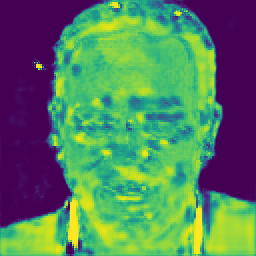}&
    \includegraphics[width=0.24\linewidth]{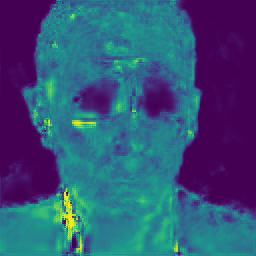}\\
    \end{tabular} 
    \vspace{0.5mm}
    \caption{IFQA results in face manipulation scenarios. Brighter area indicates the higher quality.}
    % using StarGANv2.}
    \label{fig:StarganV2}
\end{figure}

\noindent{\textbf{IFQA in face manipulation tasks:}} 
% Face domain is a field that has been actively studied for its geometrical characteristics, ease of acquisition, and various baseline techniques.
% For this reason, 
There are increasing attempts to use face images to demonstrate the superiority and effectiveness of the methodology in general image generation or image-to-image translation tasks, \eg StarGANv2~\cite{choi2020stargan} and U-GAT-IT~\cite{kim2019u}.
These tasks commonly use conventional metrics such as FID~\cite{FID} and LPIPS~\cite{LPIPS} to evaluate results.
Figure~\ref{fig:StarganV2} shows the image-to-image translation results using StarGANv2 and their visualization maps.
Overall results especially for `Output C' show the possibility that our metric designed to assess face restoration models can also be used to evaluate the results of various face-related image generation tasks.

\noindent{\textbf{IFQA as objective function:}} 
To evaluate the generalization ability of IFQA, we adopted it as an additional objective function in StarGAN v2, which is the state-of-the-art method for face manipulation. The proposed IFQA metric evaluates the per-pixel realness of generated images and gives feedback to the generator during the training phase. As a result, the generator is not only trained to generate diverse styles of images but also trained to generate image realness. Table~\ref{tab:FRQA_as_object_function} clearly shows that our IFQA strategies enhance the performance of StarGAN v2 in terms of FID. 
\begin{table}[]
\small
    \centering
\caption{Quantitative StarGAN v2 performance comparison with and without our metric as an additional objective function.
}
    \begin{tabular}{ l | c c }\hline
    \toprule
    \multirow{2}{*}{Method} & \multicolumn{2}{c}{FID$\downarrow$ }\\\cline{2-3} %
    & \multicolumn{1}{l|}{Latent-guided} & Reference-guided \\\hline%\cline{1-3}
    StarGAN v2 w/o IFQA & 14.6657& 23.7138 \\
    StarGAN v2 w IFQA & \textbf{13.8008}& \textbf{22.6457} \\
    \bottomrule
    \end{tabular} 
    \label{tab:FRQA_as_object_function}
\end{table}

\section{Conclusion}
The face domain cannot be seen merely as a common object category because of facial geometry, variety of applications, and psychological evidence.
Nevertheless, existing face image restoration studies have used general IQA metrics.
% This study introduces a novel face-centric metric based on two sub-networks. One is a face restoration model, while the other is a per-pixel discriminator that provides interpretable quality scores.
The main finding of our study is tiny distortions in facial primary regions have a significant impact on human perception. Considering this, our framework arbitrarily swaps primary regions among low-quality, high-quality, and restored images and utilizes them as supervision for the discriminator. As a result, IFQA metric shows higher correlations with human visual perception than traditional general metrics across various architectures and scenarios.

\section*{Acknowledgement}
\vspace{-1.0mm}
This work was partly supported by the National Research Foundation of Korea (NRF) grant funded by the Korea government (MSIT) (No. NRF-2021R1F1A1054569, No. 2022R1A4A1033549).
This work was partly supported by Institute of Information \& communications Technology Planning \& Evaluation (IITP) grant funded by the Korea government (MSIT) (No. RS-2022-00155915, Artificial Intelligence Convergence Innovation Human Resources Development (Inha University)).
We thank Eunkyung Jo for helpful feedback on human study design and Jaejun Yoo for constructive comments on various experimental protocols.

\appendix
\section{Implementation Details}
Our framework for IFQA consists of two sub-networks where one is a face image restoration model while the other is a per-pixel discriminator. We describe the model architecture and implementation details to facilitate the reproduction of our proposed metric results.

\subsection{Generator for image restoration}
A simple U-Net-based \cite{UNetGAN} generator model is used to train the IFQA framework. 
The generator is trained to generate 256 $\times$ 256 high-quality (HQ) images from the low-quality (LQ) images and to fool the discriminator. This generator has an encoder-decoder architecture consisting of five down-sampling and six up-sampling residual blocks. We use the average pooling in down-sampling blocks and nearest-neighbor interpolation in up-sampling blocks. We use hyperbolic tangent as a final output activation. The detailed architecture of the generator is provided in Table~\ref{tab:Generator_Arc}. Here, the kernel size, stride, and padding factors of the convolutional operations are indicated as indicate $k$, $s$, and $p$, respectively.

\subsection{Discriminator for quality assessment}
Our discriminator network takes a mixed image by facial primary region swap (FPRS) as an input and outputs a pixel-wise probability map with the exact spatial resolution as the input. 
This model also has an encoder-decoder architecture with skip-connection. 
The encoder is VGG-19~\cite{VGG16}-based architecture. We split VGG-19 into four blocks. Each block contains as follows: 1) Encoder block1: VGG-19 block1 and 2, 2) Encoder block2: VGG-19 block3, 3) Encoder block3: VGG-19 block4, 4) Encoder block4: VGG-19 block5.
% We set each layer before the max pooling operation and do not use the last fully connected layer. 
The decoder consists of four up-sampling residual blocks as same as a generator. 
We use sigmoid function as an output activation to produce each pixel value from zero to one. 
Overall details of discriminator architecture are shown in Table~\ref{tab:Discriminator_Arc}.
% We also present the setting for the main hyperparameters used to train IFQA framework in Table~\ref{tab:hyparm}.

\begin{table}[]
\centering
\begin{tabular}{lcc}\hline
\toprule
Layer& Filter& Output size \\\hline
Input Image & -& (256, 256, 3)\\\hline
Encoder block1& $k$, $s$, $p$ = \{3, 1, 1\}& (128, 128, 64)  \\ %\hline
Encoder block2& $k$, $s$, $p$ = \{3, 1, 1\}& (64, 64, 128)  \\ %\hline
Encoder block3& $k$, $s$, $p$ = \{3, 1, 1\}& (32, 32, 256)  \\ %\hline
Encoder block4& $k$, $s$, $p$ = \{3, 1, 1\}& (16, 16, 512)  \\ %\hline
Encoder block5& $k$, $s$, $p$ = \{3, 1, 1\}& (8, 8, 1024)  \\ \hline
Decoder block1& $k$, $s$, $p$ = \{3, 1, 1\}& (16, 16, 512)  \\ %\hline
Decoder block2& $k$, $s$, $p$ = \{3, 1, 1\}& (32, 32, 256)  \\ %\hline
Decoder block3& $k$, $s$, $p$ = \{3, 1, 1\}& (64, 64, 128)  \\ %\hline
Decoder block4& $k$, $s$, $p$ = \{3, 1, 1\}& (128, 128, 64)  \\  
Decoder block5& $k$, $s$, $p$ = \{3, 1, 1\}& (256, 256, 64)  \\ 
Decoder block6& $k$, $s$, $p$ = \{3, 1, 1\}& (256, 256, 3)  \\\hline 
Tanh& -& (256, 256, 3) \\ \bottomrule
\end{tabular}
\caption{Model architecture of the generator.}
\vspace{-3mm}
\label{tab:Generator_Arc}
\end{table}

\begin{table}[]
\centering
\begin{tabular}{lcc}\hline
\toprule
Layer& Filter& Output size \\\hline
Input Image & -& (256, 256, 3)\\\hline
Encoder block1& $k$, $s$, $p$ = \{3, 1, 1\}& (128, 128, 128)  \\ %\hline
Encoder block2& $k$, $s$, $p$ = \{3, 1, 1\}& (64, 64, 256)  \\ %\hline
Encoder block3& $k$, $s$, $p$ = \{3, 1, 1\}& (32, 32, 512)  \\ %\hline
Encoder block4& $k$, $s$, $p$ = \{3, 1, 1\}& (16, 16, 512)  \\ \hline
Decoder block1& $k$, $s$, $p$ = \{3, 1, 1\}& (32, 32, 512)  \\ %\hline
Decoder block1& $k$, $s$, $p$ = \{3, 1, 1\}& (64, 64, 256)  \\ %\hline
Decoder block1& $k$, $s$, $p$ = \{3, 1, 1\}& (128, 128, 128)  \\  
Decoder block1& $k$, $s$, $p$ = \{3, 1, 1\}& (256, 256, 64)  \\ 
Decoder block1& $k$, $s$, $p$ = \{3, 1, 1\}& (256, 256, 1)  \\\hline
Sigmoid& & (256, 256, 1) \\ \bottomrule
\end{tabular}
\caption{Model architecture of the per-pixel discriminator.}
\vspace{-3mm}
\label{tab:Discriminator_Arc}
\end{table}

\section{Human Study Protocol}
% We carefully designed the image quality assessment survey to estimate human visual judgments for restored face images. 
Firstly, we prepared 200 face images samples. 
Each sample involves six images, consisting of LQ images and restored results from SISR models and FIR models. 
We constructed two types of test sets.
% considering full-reference (FR) and no-reference (NR) scenarios. 
We combined CelebA-HQ \cite{CelebAHQ} with FFHQ \cite{StyleGANv1_FFHQ} for the full-reference (FR) scenario and used IWF \cite{GPEN} for the no-reference (NR) scenario.
As a result, 200 samples were randomly selected from the FR and NR sets for our survey. 
Note that the number of images from the FR set is 100 and that of the NR set is 100.
To prevent participants from being biased toward ground truth (GT) images (\ie high-quality reference images), we excluded GT images from the FR set in our survey.
We also shuffled the display order of the images for each sample in the survey question.
Overall, the total number of images for the human study is 1,200.

\begin{figure*}[h]
\begin{center}   
\resizebox{0.6\textwidth}{!}{\begin{tabular}{c}
\includegraphics[ width=0.61\linewidth]{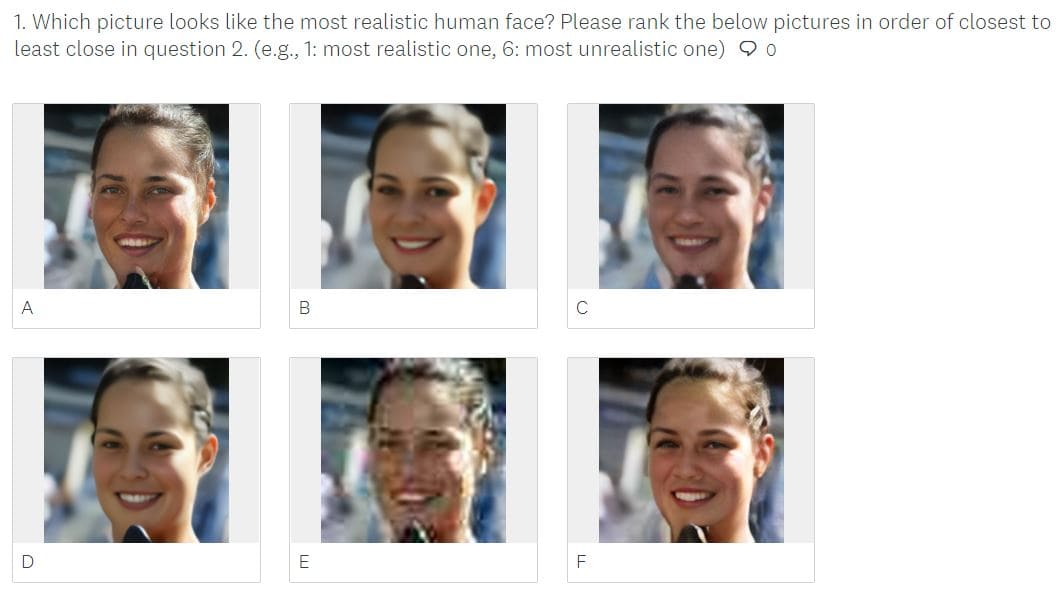}\\
\includegraphics[ width=0.61\linewidth]{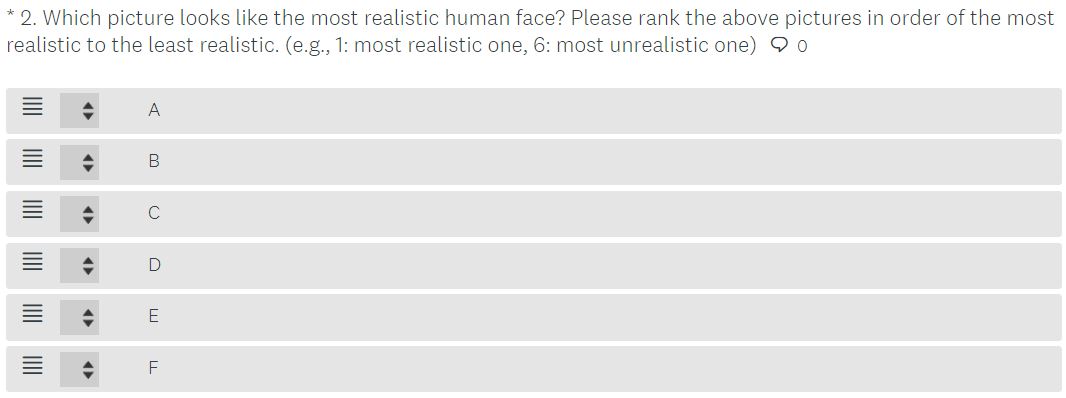}\\
\includegraphics[ width=0.58\linewidth]{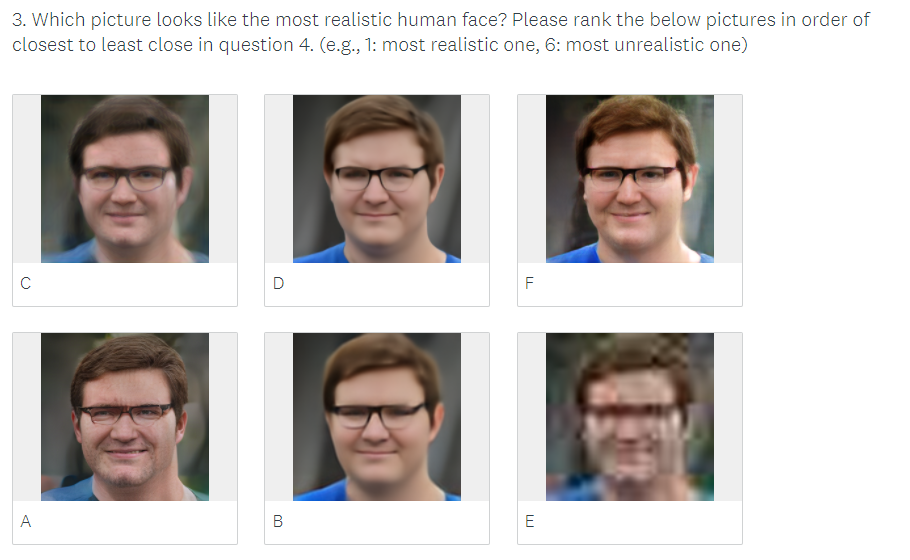}\\
\includegraphics[ width=0.62\linewidth]{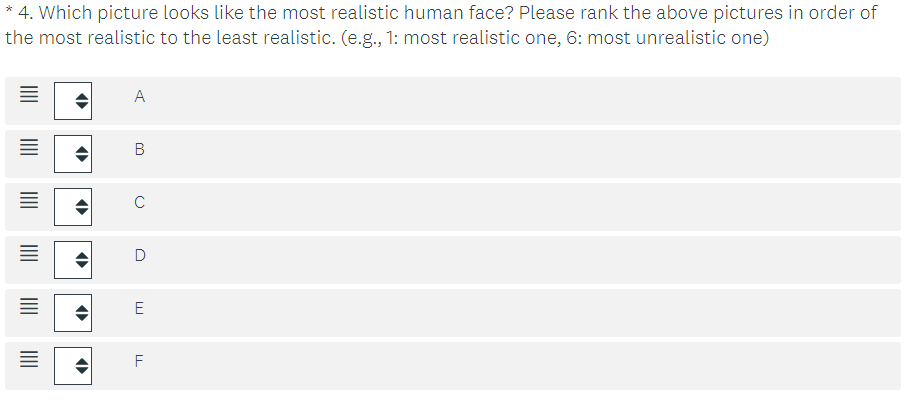}\\
\end{tabular}
}
\end{center}
\caption{Survey examples. We ask participants to rank in order of the most realistic face images among the given samples. }
\label{fig:survey_example1}
\end{figure*}

\begin{figure*}
    \centering
    \includegraphics[width=0.84\linewidth]{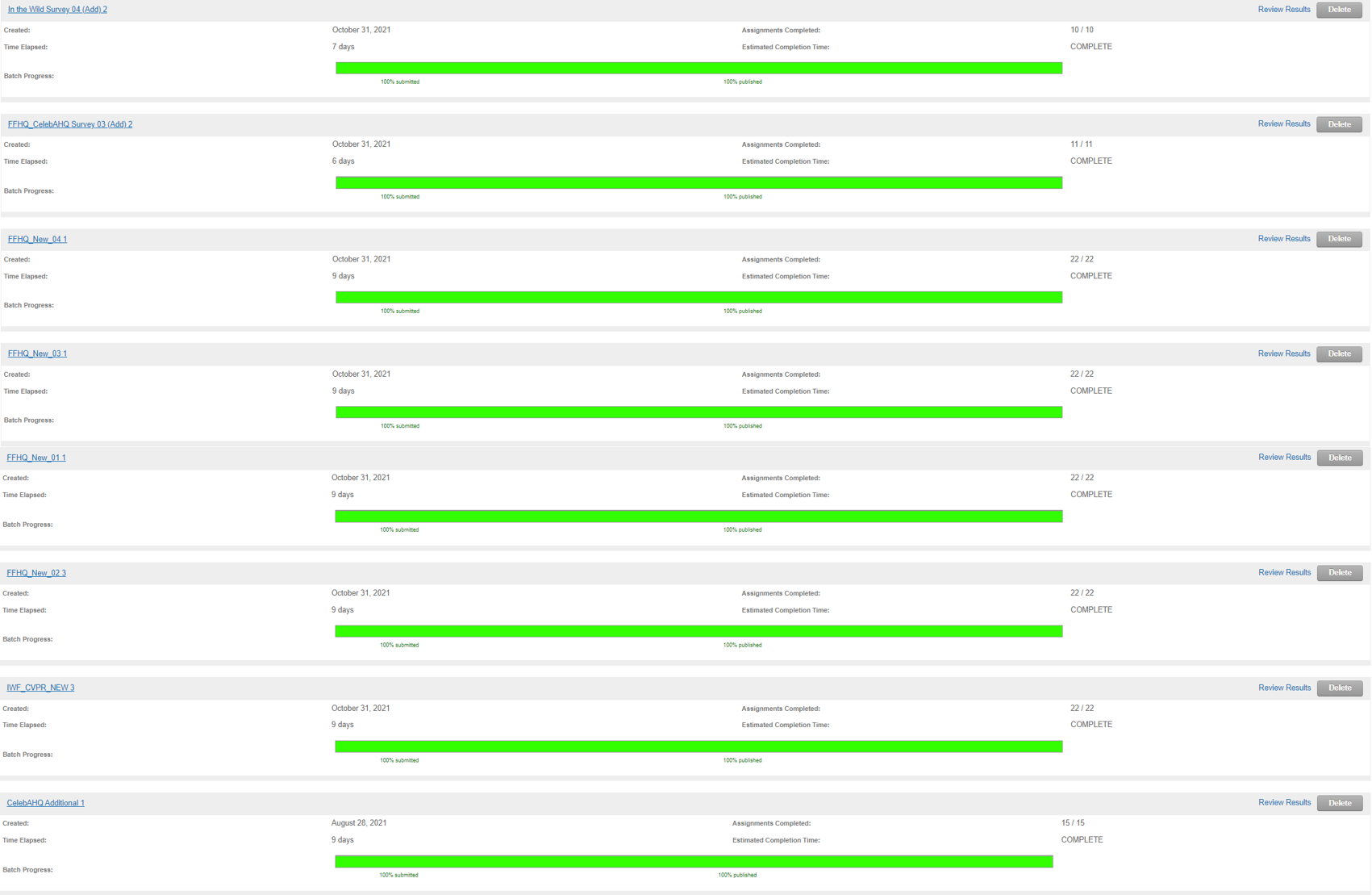}\\
    \includegraphics[width=0.84\linewidth]{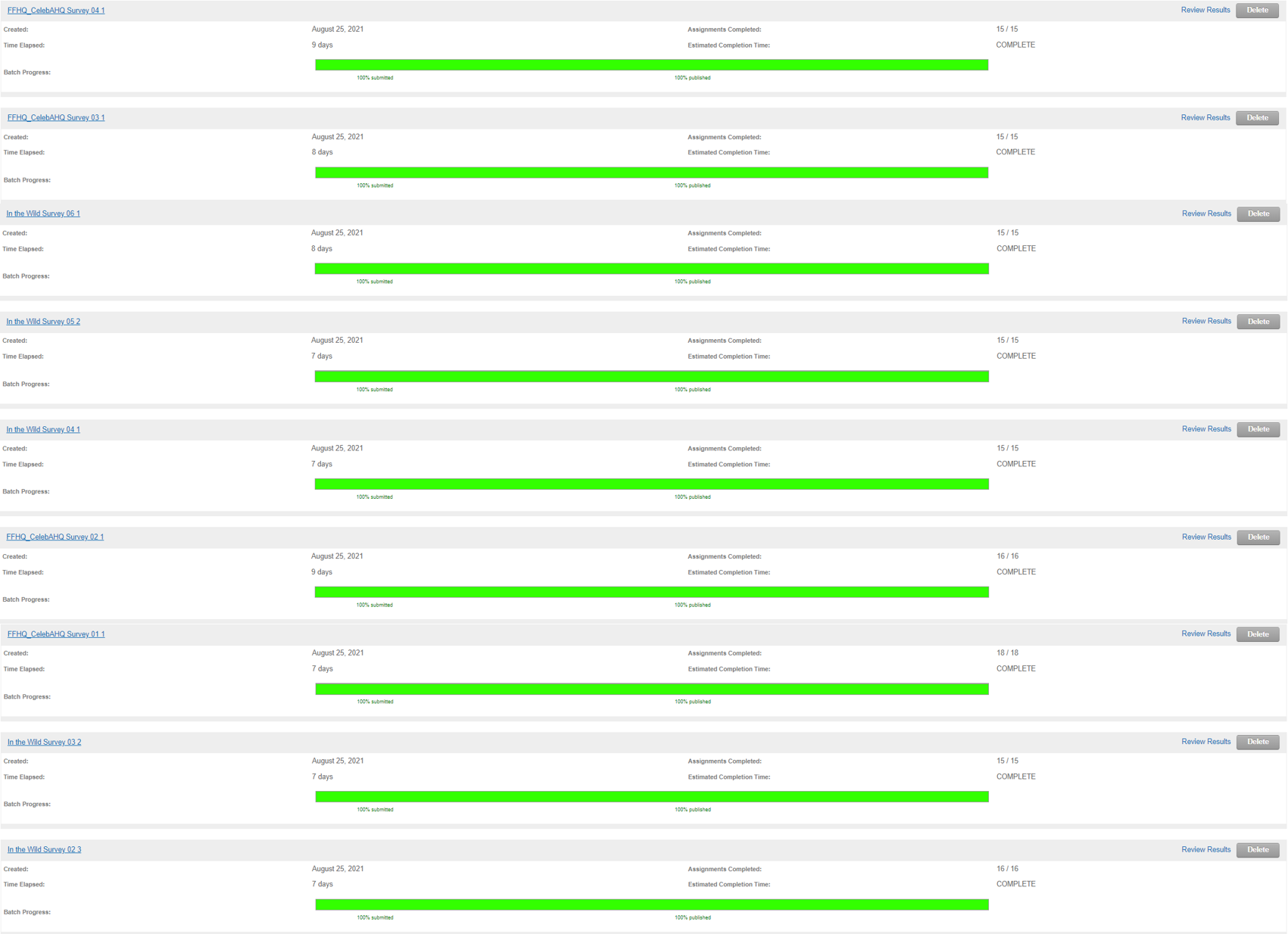}
    \caption{Survey status menu in Amazon Mechanical Turk.}
    \label{fig:AMT}
\end{figure*}

To gather the human responses systematically, we first created a survey using the survey software tool~\cite{surveymonkeyquestion} (see Figure~\ref{fig:survey_example1}) and integrated the survey with Amazon Mechanical Turk (AMT) for crowdsourcing (see Figure~\ref{fig:AMT}). 
We asked participants from AMT to rank given samples from closest to furthest to a realistic human face.
We also received responses from researchers who are majors in various AI-related fields and are not directly related to this study.
We assigned 30 subjects per sample, and the total number of responses across whole samples was 6,000.
{The final rank of each sample is calculated by the weighted average scores~\cite{surveymonkeyquestion} as follows:
\vspace{-0.1cm}
\begin{equation}
    score = \frac{1}{N}\sum_{i=1}^{6}{w_{i} r_{i}},
\end{equation}
\vspace{-0.05cm}
where $N$ refers to the number of total raters, $r_{i}$ represents the rank for $i$-th image \{1, 2, 3, 4, 5, 6\}, and $w_{i}$ is corresponding weights \{6, 5, 4, 3, 2, 1\}. As a result, each image is ranked in descending order of the calculated scores.}

\section{Image Restoration Model Training}\label{sec:2}
For quantitative and qualitative comparisons between the existing metrics and the proposed metrics, we utilize various image restoration (IR) models, including single image super-resolution methods (\eg, RCAN~\cite{RCAN}, DBPN~\cite{DBPN}) and face image restoration (FIR) models (\eg, DFDNet~\cite{Li_DFDNet}, HiFaceGAN~\cite{Yang_HiFaceGAN}, GPEN~\cite{GPEN}). 
We use the same degradation process as well as the same dataset during the IFQA framework training phase.
We train all image restoration models using the official training code and settings.

\subsection{General image restoration (GIR) models}
Since the super-resolution models aim to reconstruct low-resolution (LR) images to high-resolution (HR) images, we deliberately corrupt and down-sample HQ images to 32 $\times$ 32 resolution during both the training and inference phase. 
We then train the image restoration models to reconstruct 32 $\times$ 32 LQ images to 256 $\times$ 256 HQ images.

\vspace{-0.2cm}
\subsection{Face image restoration (FIR) models} 
FIR models aim to restore the arbitrary LQ face images to the corresponding HQ images. 
Since our metric assesses the 256 $\times$ 256 face images, we set the target resolution equal to our metric for training FIR models. Since DFDNet did not release the training codes, we train DFDNet using the same protocol as much as possible. 
% {Due to the unstable training with adversarial loss in DFDNet, we apply the pixel-wise $L2$ loss.} 
GPEN provides the pre-trained model, which restores LQ images to 512 $\times$ 512 HQ images. To hand this issue, we restore LQ images using the pre-trained model and then down-sample the restored images to 256 $\times$ 256.

\section{Qualitative results} 
\begin{figure*}[h]
    \centering
    \begin{tabular}{c@{\hskip 1pt}c@{\hskip 1pt}c@{\hskip 1pt}c@{\hskip 1pt}c@{\hskip 1pt}c@{\hskip 1pt}c}
    LQ& RCAN& DBPN& DFDNet& HiFaceGAN& GPEN& Reference\\
    \includegraphics[width=0.12\linewidth]{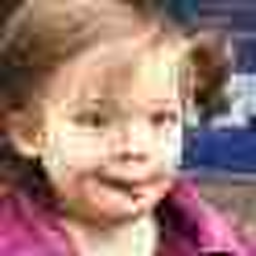}&
    \includegraphics[width=0.12\linewidth]{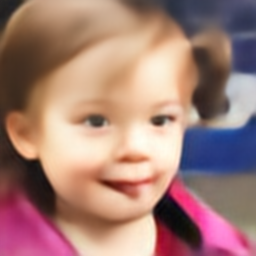}&
    \includegraphics[width=0.12\linewidth]{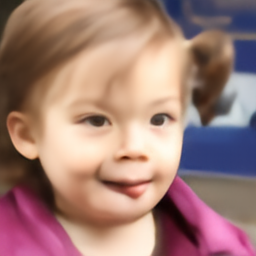}&
    \includegraphics[width=0.12\linewidth]{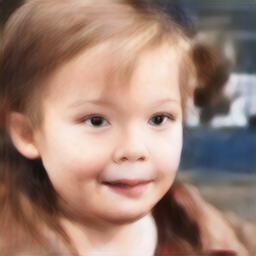}&
    \includegraphics[width=0.12\linewidth]{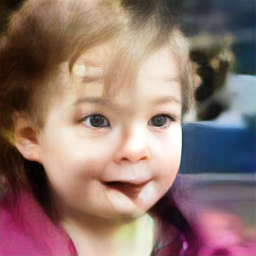}&
    \includegraphics[width=0.12\linewidth]{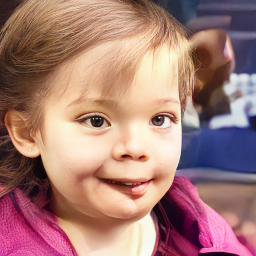}& 
    \includegraphics[width=0.12\linewidth]{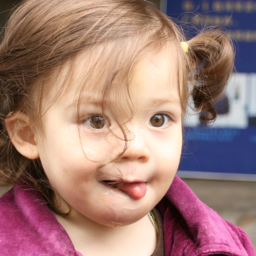}\\
    \includegraphics[width=0.12\linewidth]{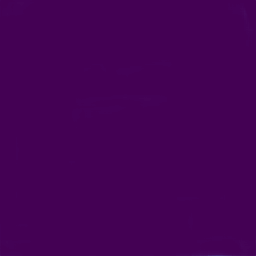}&
    \includegraphics[width=0.12\linewidth]{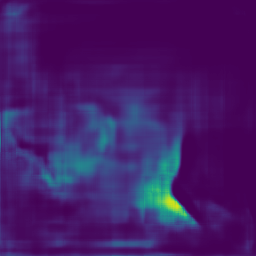}&
    \includegraphics[width=0.12\linewidth]{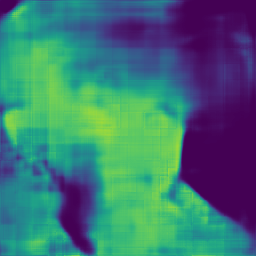}&
    \includegraphics[width=0.12\linewidth]{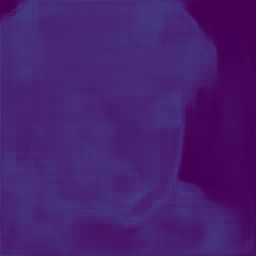}&
    \includegraphics[width=0.12\linewidth]{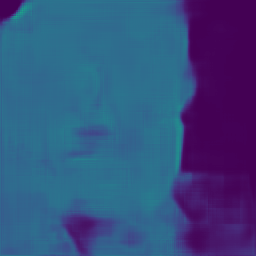}&
    \includegraphics[width=0.12\linewidth]{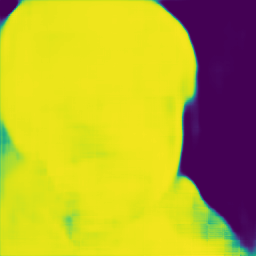}& \vspace{-1mm}
    \includegraphics[width=0.12\linewidth]{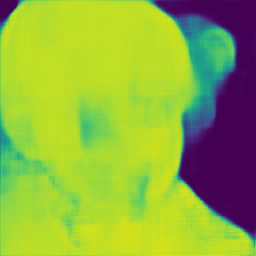}\\ 
    \small{0.0015}& \small{0.0849}& \small{0.4144}& \small{0.0888}& \small{0.2526}& \small{0.7301}& \small{0.7003}\\ \vspace{-4mm}
    \\ \midrule
    \includegraphics[width=0.12\linewidth]{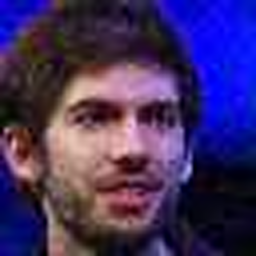}&
    \includegraphics[width=0.12\linewidth]{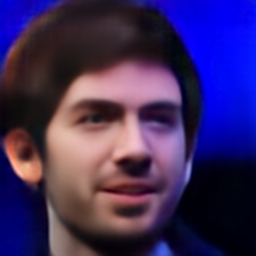}&
    \includegraphics[width=0.12\linewidth]{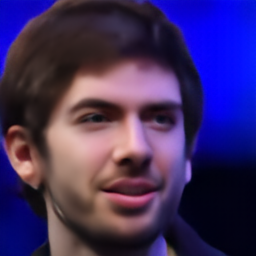}&
    \includegraphics[width=0.12\linewidth]{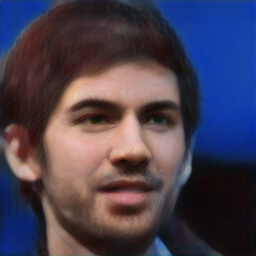}&
    \includegraphics[width=0.12\linewidth]{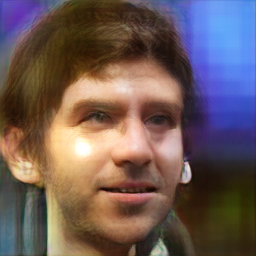}&
    \includegraphics[width=0.12\linewidth]{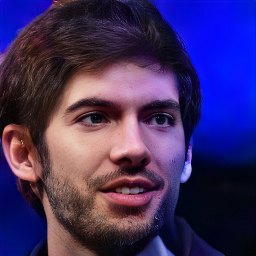}& 
    \includegraphics[width=0.12\linewidth]{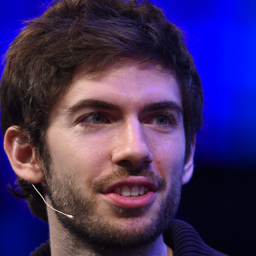}\\ 
    \includegraphics[width=0.12\linewidth]{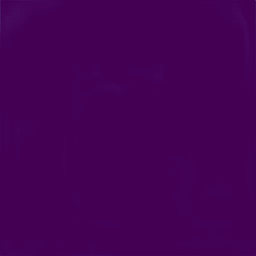}&
    \includegraphics[width=0.12\linewidth]{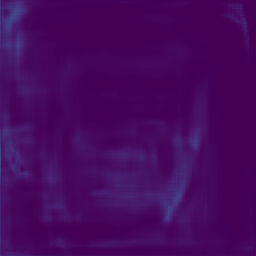}&
    \includegraphics[width=0.12\linewidth]{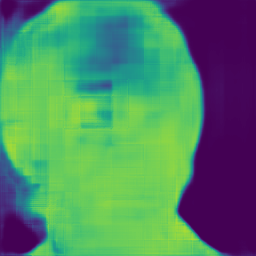}&
    \includegraphics[width=0.12\linewidth]{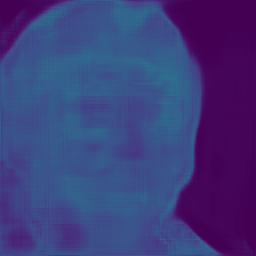}&
    \includegraphics[width=0.12\linewidth]{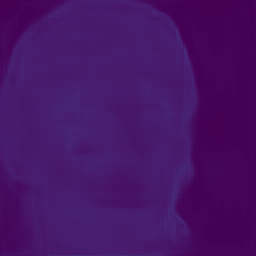}&
    \includegraphics[width=0.12\linewidth]{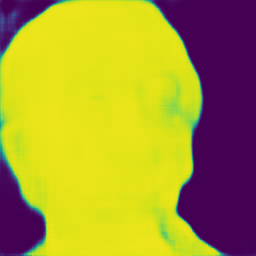}& \vspace{-1mm}
    \includegraphics[width=0.12\linewidth]{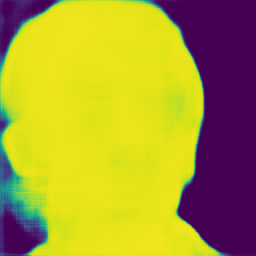}\\
    \small{0.0024}& \small{0.0238}& \small{0.4979}& \small{0.1853}& \small{0.0524}& \small{0.6443}& \small{0.6642}\\\vspace{-4mm}
    \\ \midrule
    \includegraphics[width=0.12\linewidth]{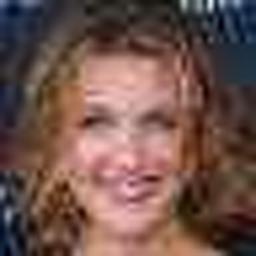}&
    \includegraphics[width=0.12\linewidth]{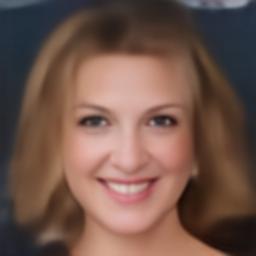}&
    \includegraphics[width=0.12\linewidth]{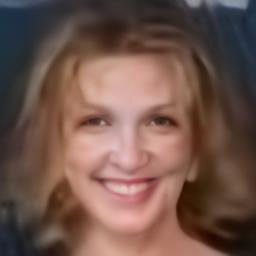}&
    \includegraphics[width=0.12\linewidth]{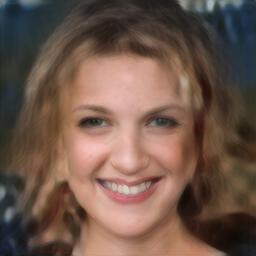}&
    \includegraphics[width=0.12\linewidth]{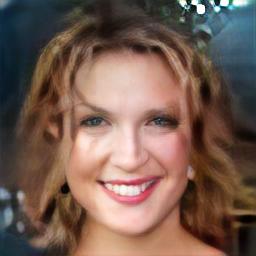}&
    \includegraphics[width=0.12\linewidth]{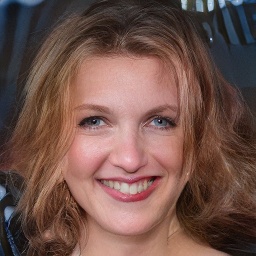}& 
    \includegraphics[width=0.12\linewidth]{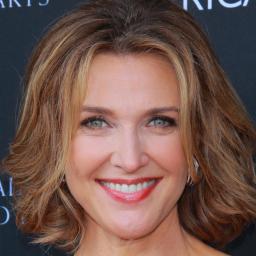}\\
    \includegraphics[width=0.12\linewidth]{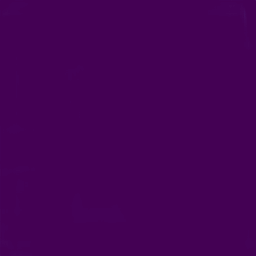}&
    \includegraphics[width=0.12\linewidth]{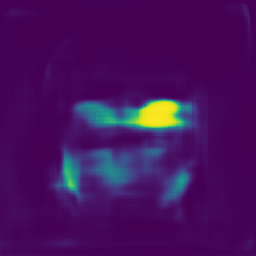}&
    \includegraphics[width=0.12\linewidth]{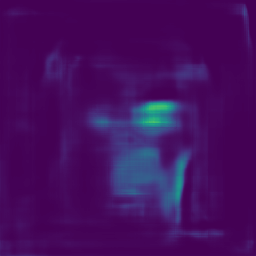}&
    \includegraphics[width=0.12\linewidth]{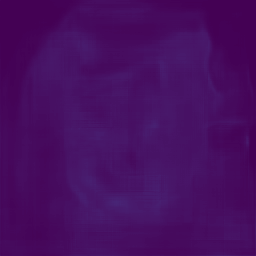}&
    \includegraphics[width=0.12\linewidth]{Fig_Qualitative/00079/Viz/DBPN_0.0423.png}&
    \includegraphics[width=0.12\linewidth]{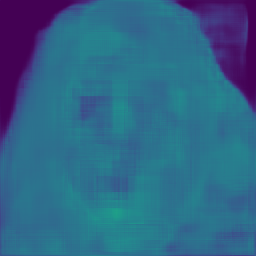}& \vspace{-1mm}
    \includegraphics[width=0.12\linewidth]{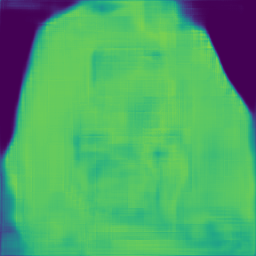}\\
    \small{0.0012}& \small{0.0566}& \small{0.0423}& \small{0.0358}& \small{0.0423}& \small{0.3351}& \small{0.6212}\\\vspace{-4mm}
       \\ \midrule 
    \includegraphics[width=0.12\linewidth]{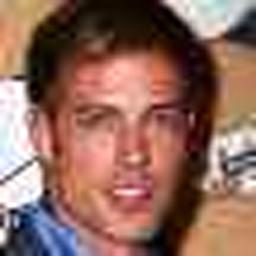}&
    \includegraphics[width=0.12\linewidth]{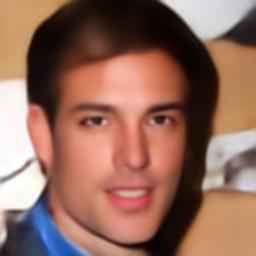}&
    \includegraphics[width=0.12\linewidth]{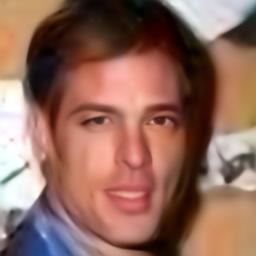}&
    \includegraphics[width=0.12\linewidth]{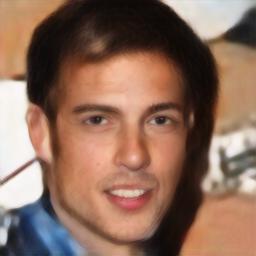}&
    \includegraphics[width=0.12\linewidth]{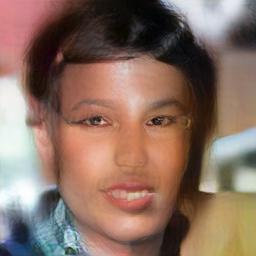}&
    \includegraphics[width=0.12\linewidth]{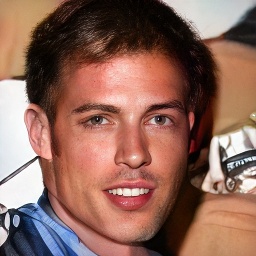}& 
    \includegraphics[width=0.12\linewidth]{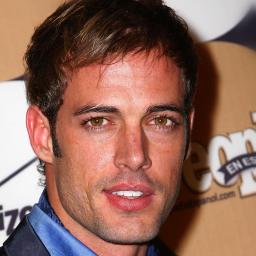}\\ 
    \includegraphics[width=0.12\linewidth]{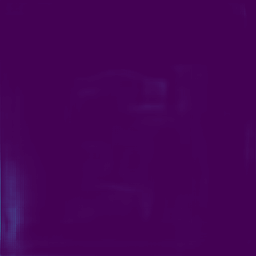}&
    \includegraphics[width=0.12\linewidth]{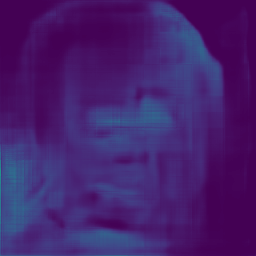}&
    \includegraphics[width=0.12\linewidth]{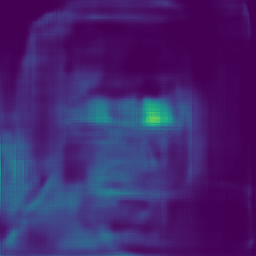}&
    \includegraphics[width=0.12\linewidth]{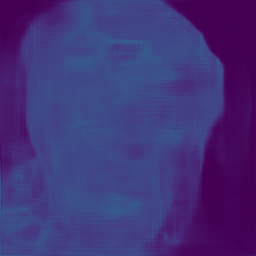}&
    \includegraphics[width=0.12\linewidth]{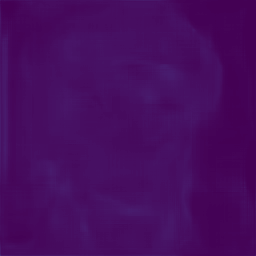}&
    \includegraphics[width=0.12\linewidth]{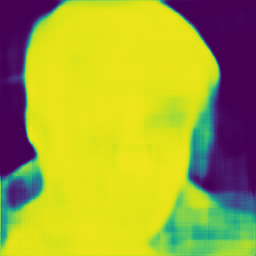}& \vspace{-1mm}
    \includegraphics[width=0.12\linewidth]{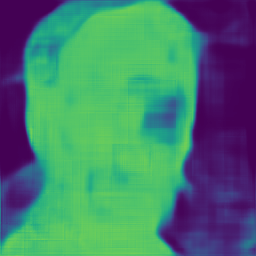}\\
    \small{0.0056}& \small{0.0926}& \small{0.1108}& \small{0.1515}& \small{0.0309}& \small{0.6951}& \small{0.4496}\\ \vspace{-4mm}
    \end{tabular}
    \caption{Additional evaluation results (\ie pixel-level and image-level scores) of our IFQA metric on FFHQ and CelebA-HQ.}
    \label{fig:PFQE_Quali_FR}
\end{figure*}

\begin{figure*}[h]
    \centering
    \begin{tabular}{c@{\hskip 1pt}c@{\hskip 1pt}c@{\hskip 1pt}c@{\hskip 1pt}c@{\hskip 1pt}c}
    LQ& RCAN& DBPN& DFDNet& HiFaceGAN& GPEN\\
    \includegraphics[width=0.13\linewidth]{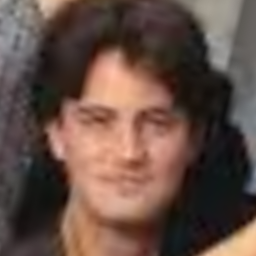}&
    \includegraphics[width=0.13\linewidth]{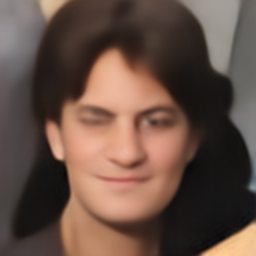}&
    \includegraphics[width=0.13\linewidth]{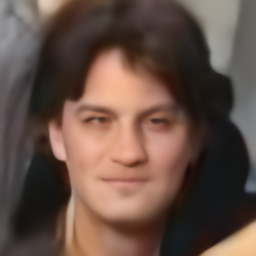}&
    \includegraphics[width=0.13\linewidth]{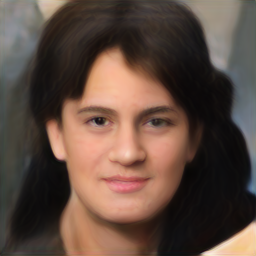}&
    \includegraphics[width=0.13\linewidth]{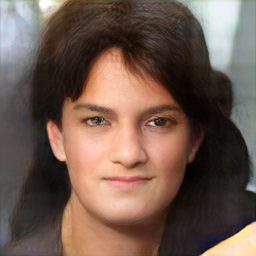}&
    \includegraphics[width=0.13\linewidth]{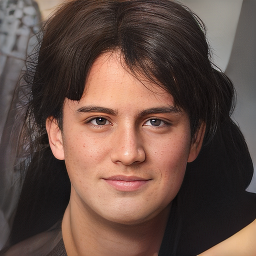}\\
    \includegraphics[width=0.13\linewidth]{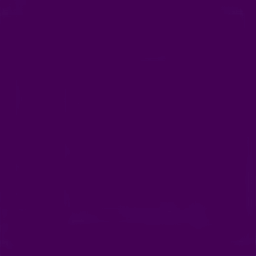}&
    \includegraphics[width=0.13\linewidth]{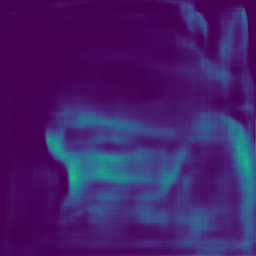}&
    \includegraphics[width=0.13\linewidth]{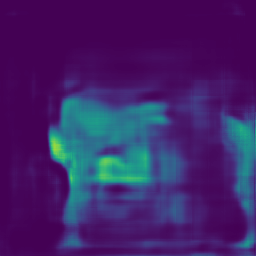}&
    \includegraphics[width=0.13\linewidth]{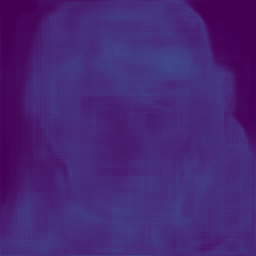}&
    \includegraphics[width=0.13\linewidth]{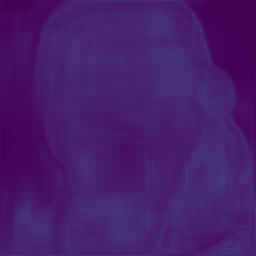}& \vspace{-1mm}
    \includegraphics[width=0.13\linewidth]{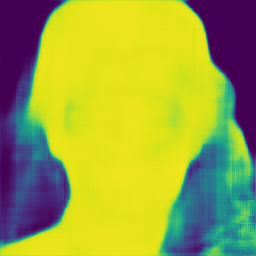}\\
    \small{0.0015}& \small{0.0948}& \small{0.1101}& \small{0.1075}& \small{0.0802}& \small{0.6673}\\ \vspace{-4mm}
    \\ \midrule 
    \includegraphics[width=0.13\linewidth]{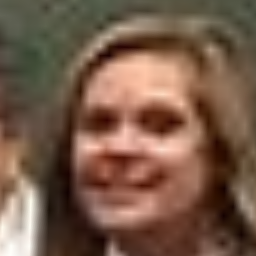}&
    \includegraphics[width=0.13\linewidth]{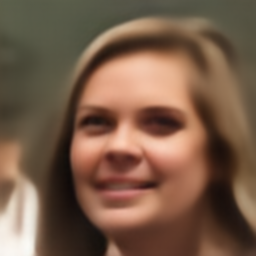}&
    \includegraphics[width=0.13\linewidth]{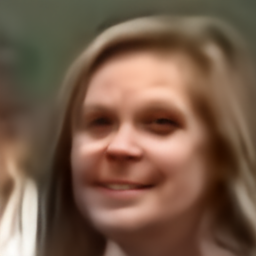}&
    \includegraphics[width=0.13\linewidth]{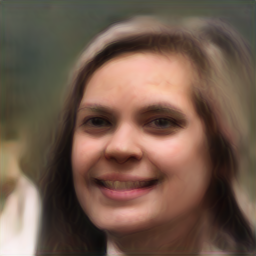}&
    \includegraphics[width=0.13\linewidth]{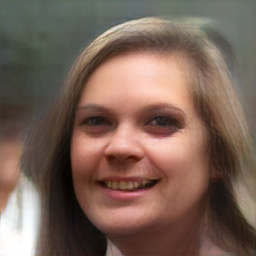}&
    \includegraphics[width=0.13\linewidth]{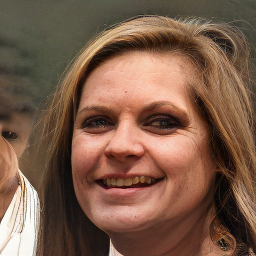}\\
    \includegraphics[width=0.13\linewidth]{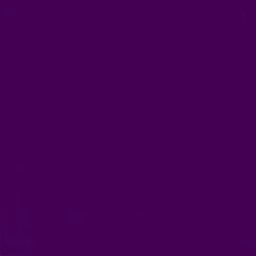}&
    \includegraphics[width=0.13\linewidth]{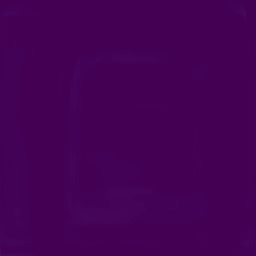}&
    \includegraphics[width=0.13\linewidth]{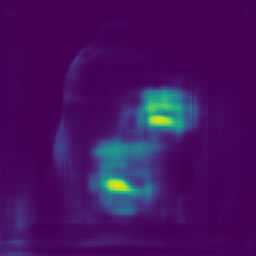}&
    \includegraphics[width=0.13\linewidth]{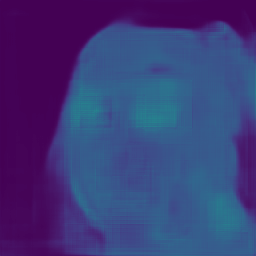}&
    \includegraphics[width=0.13\linewidth]{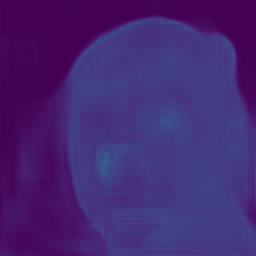}& \vspace{-1mm}
    \includegraphics[width=0.13\linewidth]{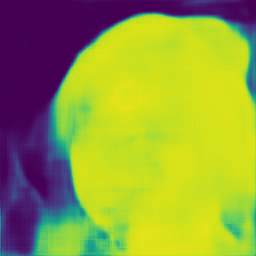}\\
    \small{0.001}& \small{0.0029}& \small{0.0588}& \small{0.1824}& \small{0.1117}& \small{0.6349}\\\vspace{-4mm}
    \\ \midrule 
        \includegraphics[width=0.13\linewidth]{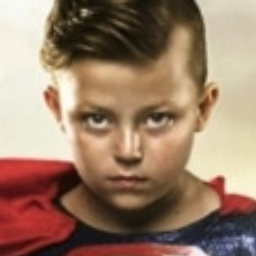}&
    \includegraphics[width=0.13\linewidth]{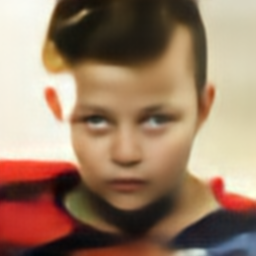}&
    \includegraphics[width=0.13\linewidth]{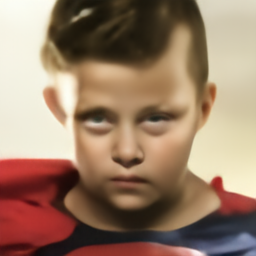}&
    \includegraphics[width=0.13\linewidth]{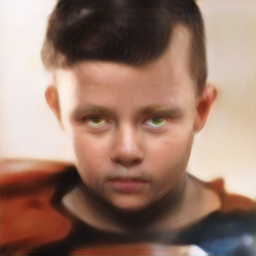}&
    \includegraphics[width=0.13\linewidth]{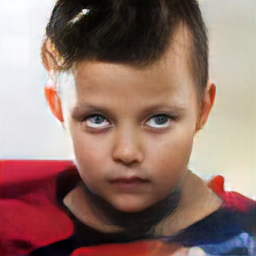}&
    \includegraphics[width=0.13\linewidth]{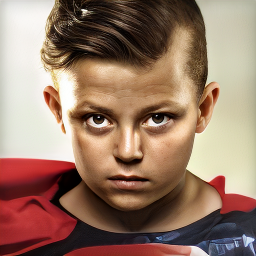}\\
    \includegraphics[width=0.13\linewidth]{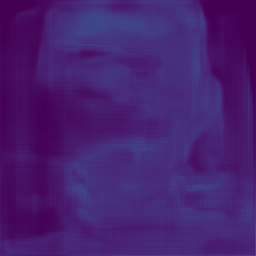}&
    \includegraphics[width=0.13\linewidth]{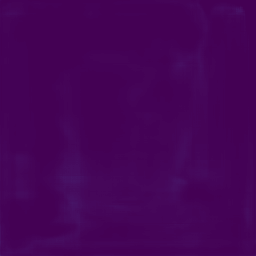}&
    \includegraphics[width=0.13\linewidth]{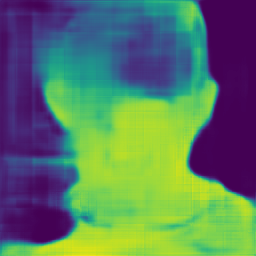}&
    \includegraphics[width=0.13\linewidth]{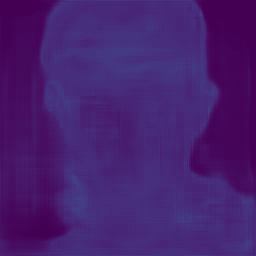}&
    \includegraphics[width=0.13\linewidth]{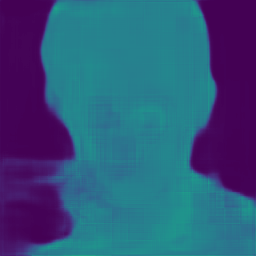}& \vspace{-1mm}
    \includegraphics[width=0.13\linewidth]{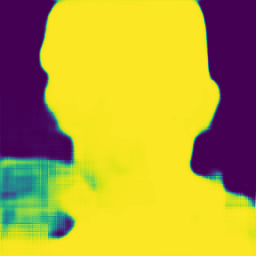}\\
    \small{0.0905}& \small{0.0082}& \small{0.4454}& \small{0.0888}& \small{0.2823}& \small{0.6981}\\\vspace{-4mm}
    \\ \midrule 
        \includegraphics[width=0.13\linewidth]{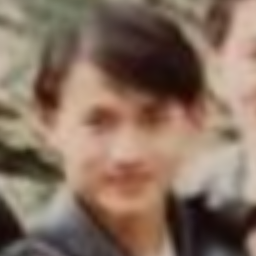}&
    \includegraphics[width=0.13\linewidth]{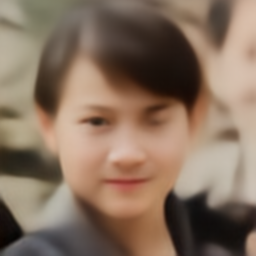}&
    \includegraphics[width=0.13\linewidth]{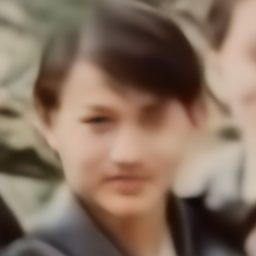}&
    \includegraphics[width=0.13\linewidth]{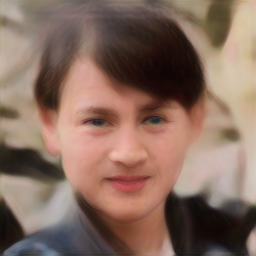}&
    \includegraphics[width=0.13\linewidth]{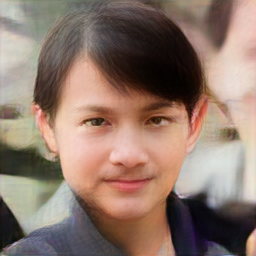}&
    \includegraphics[width=0.13\linewidth]{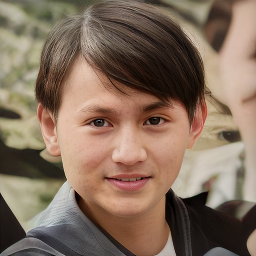}\\
    \includegraphics[width=0.13\linewidth]{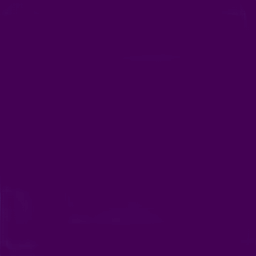}&
    \includegraphics[width=0.13\linewidth]{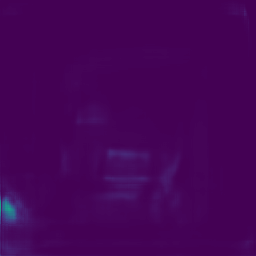}&
    \includegraphics[width=0.13\linewidth]{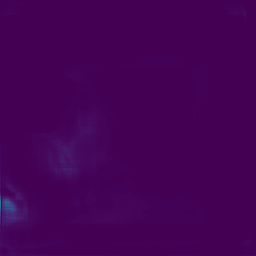}&
    \includegraphics[width=0.13\linewidth]{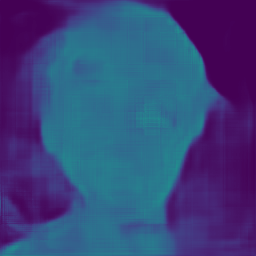}&
    \includegraphics[width=0.13\linewidth]{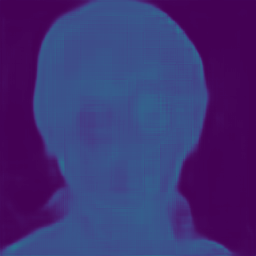}& \vspace{-1mm}
    \includegraphics[width=0.13\linewidth]{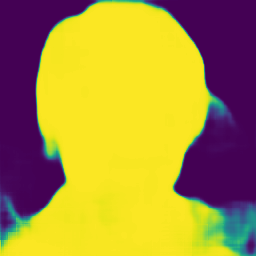}\\
    \small{0.0011}& \small{0.0090}& \small{0.0050}& \small{0.2219}& \small{0.1464}& \small{0.6127}\\ \vspace{-4mm}
    \end{tabular}
    \caption{Additional evaluation results (\ie pixel-level and image-level scores) of our IFQA metric on IWF.}
    \label{fig:qual_NR}
\end{figure*}

We extensively provide assessment results of the proposed IFQA on the FR dataset (\ie FFHQ and CelebA-HQ) and the NR dataset (\ie IWF). 
FR datasets include the LQ, references, and restored face images from RCAN, DBPN, DFDNet, HiFaceGAN, and GPEN.
Since the no-reference IWF dataset does not have corresponding HQ references images, we show LQ and restored images. 
We can see that the proposed metric has a high correlation with human judgment from each image and the corresponding pixel-level and image-level scores in Figure~\ref{fig:PFQE_Quali_FR} and~\ref{fig:qual_NR}.

\clearpage
 \clearpage

\clearpage 

\clearpage 
 
{\small
\bibliographystyle{ieee_fullname}
\bibliography{egbib}
}

\end{document}